\newif\ifssltr

\ssltrtrue

\documentclass[11pt]{article}

\ifssltr
  \usepackage[report]{ssl-report}
  \usepackage{amssymb,amsthm}
  \usepackage{booktabs,tabularx,placeins}
  \usepackage[numbers,sort&compress]{natbib}
  \usepackage[colorlinks=true,linkcolor=sslblue,citecolor=sslblue,urlcolor=sslblue]{hyperref}
\else
  \usepackage[T1]{fontenc}
  \usepackage[margin=1in]{geometry}
  \usepackage{amsmath,amssymb,amsthm}
  \usepackage{fourier}
  \usepackage{microtype}
  \usepackage{booktabs,tabularx,placeins}
  \usepackage[hidelinks]{hyperref}
\fi

\newtheorem{theorem}{Theorem}
\newtheorem{proposition}{Proposition}
\newtheorem{corollary}{Corollary}
\theoremstyle{definition}
\newtheorem{definition}{Definition}

\newcommand{\N}{\mathbb{N}}
\newcommand{\Str}{\Sigma^{*}}                 
\newcommand{\Traces}{A^{\le\omega}}           
\newcommand{\tr}{\operatorname{tr}}           
\newcommand{\Out}{\operatorname{Out}}         
\newcommand{\Harm}{\mathrm{HARM}}
\newcommand{\Bad}{\mathrm{Bad}}
\newcommand{\Sent}{\mathrm{Sent}}
\newcommand{\Th}{\mathrm{Th}(\N)}
\newcommand{\halts}{{\downarrow}}
\newcommand{\diverges}{{\uparrow}}
\newcommand{\mred}{\le_{\mathrm{m}}}          
\newcommand{\co}[1]{\overline{#1}}            
\newcommand{\symdiff}{\mathbin{\triangle}}    
\newcommand{\emptystr}{\varepsilon}
\newcommand{\dlv}{\mathord{!}}                

\newcommand{\paperabstract}{%
Expecting guardrails to provide a general guarantee of safety fails to
recognize the limits of both computational enforcement and policy
specification. We argue that promising such perfection is an ethical failure
of precision, and that invoking it to abdicate responsibility is a moral
failure. Guardrails can support responsible action, but cannot relieve
researchers, developers, policymakers, or users of their obligations to
examine what they claim, permit, and prevent.

We establish a precise computational boundary. If every finite string is a
possible output of a language model, a computable guard that blocks every
forbidden output and passes every permitted one exists if and only if the
policy is decidable. If the policy is undecidable, every guard misclassifies infinitely many
strings, and every guard that blocks all forbidden outputs excludes infinitely
many permitted ones. We apply this characterization to arithmetic truth,
program behavior, and harmful instructions to an idealized recipient. We
distinguish guards from execution monitors, identify bounded and interceptable
cases admitting exact enforcement, and explain why the results establish no
positive lower bound on average error.

These distinctions matter to the practical and ethical interpretation of the
proof. Exact enforcement, where possible, does not establish that a policy
adequately captures safety. Conservative exclusion cannot be presumed harmless:
exclusions may withhold warnings, remedies, or
challenges needed to recognize the policy's own defects. Human oversight
likewise cannot substitute for the exercise of judgment: an apparent
controller can become an instrument of purposes they have ceased to examine.
Responsible governance therefore requires continued critical challenge and
the intellectual and moral humility to recognize possible flaws in our
machines, rules, policies, and ourselves. The impossibility of a general
assurance of perfection is a condition responsible practice must confront,
not a responsibility that another guardrail---or another guard---can remove.}

\ifssltr
  \title{No Perfect Guardrail \& No Perfect Guard}
  \sslsubtitle{On the Decidability of Output Policies for Large Language Models}
  \author{Ahmed Amer\textsuperscript{1} \and Loren Kinczel\textsuperscript{2} \and Darrell D. E. Long\textsuperscript{1,3}}
  \sslaffiliation{\textsuperscript{1}\sslname, Department of Computer Science \& Engineering, Santa Clara University\\
    \textsuperscript{2}Cultivate Investments\qquad \textsuperscript{3}Pentexoire Consulting}
  \date{November 5, 2026}
  \sslreportnumber{SSL-TR-2026-002}
  \sslstatus{Preprint v0.1}
  \sslkeywords{decidability; output policies; large language models; guardrails; responsibility}
  \sslcitation{Ahmed Amer, Loren Kinczel, and Darrell D. E. Long. \emph{No Perfect Guardrail \& No Perfect Guard: On the Decidability of Output Policies for Large Language Models}. SSL Technical Report Series, SSL-TR-2026-002. Sustainable Systems Laboratory, Santa Clara University, November 5, 2026. Preprint v0.1. \sslreportlink}
  \sslabstract{\paperabstract}
\else
  \title{No Perfect Guardrail \& No Perfect Guard:\\ On the Decidability of Output Policies for Large Language Models}
  \author{Ahmed Amer \\ \small Santa Clara University \and
  Loren Kinczel\\ \small Cultivate Investments
          \and
          Darrell D. E. Long\\ \small Santa Clara University\\ \small Pentexoire Consulting}
  \date{}
\fi

\ifssltr
  \makeatletter
  \patchcmd{\sslcoversheet}{\@title}{No Perfect Guardrail\\\& No Perfect Guard}{}{%
    \PackageError{manuscript}{Could not set cover title line break}{}%
  }
  \makeatother
  \patchcmd{\sslcoversheet}{\restoregeometry}{%
    \par\vskip .6em{\scriptsize\color{black!70}%
      AI tools were used in the preparation of this version (v0.1) of the manuscript.\par}%
    \restoregeometry
  }{}{\PackageError{manuscript}{Could not add cover preparation disclosure}{}}
\fi

\begin{document}
\ifssltr
  \apptocmd{\sslcoversheet}{
\begingroup
\setlength{\parindent}{0pt}
\setlength{\parskip}{0.2em}
\section*{Executive Summary}
\label{sec:executive-summary}

{\large\bfseries No general promise of perfect guardrails can be fulfilled.\\
Our responsibility cannot be shed.\par}

Any general assurance of a perfect guardrail is simply an impossibility.
Particular policies can be enforced exactly, but that does not support a
general promise of perfection. We argue that these limits demand precision in
how we talk about the technology and both individual and institutional
responsibility for managing its risks. Neither promises of general perfection,
nor the impossibility of fulfilling them, can absolve any of us of that
responsibility.

\textbf{What we establish.}
We prove that, when a language model can produce any finite string, a
computable guard can block every forbidden output and pass every permissible
one if and only if an algorithm can decide whether an output violates the
policy. For policies where no such algorithm exists, better engineering
cannot deliver both guarantees, even if the policy is precisely defined and
faithfully captures its authors' intent. Blocking all forbidden outputs then
requires excluding infinitely many permissible ones. This does not tell us how
often errors occur in practice or establish that guardrails are ineffective.
It does rule out a general guarantee of perfect enforcement.

\textbf{Why blocking more does not settle safety.}
Withholding permissible information may prevent us from recognizing a danger,
finding a remedy, or challenging a defective policy. Compliance with guardrails
cannot establish that these exclusions are harmless. Nor could even perfect
enforcement establish that we defined the right policies and purposes. Simply
put, what a system prevents has consequences, not just what it permits. Even a
guardrail that prevents every unsafe output could create risk through its
exclusions.

\textbf{What we ask of policymakers.}
Responsible AI policy benefits neither from doomsaying nor from demands for
impossible safeguards. We ask those proposing systems or restrictions to
answer three questions:
\begin{enumerate}
  \setlength{\itemsep}{0.25em}
  \setlength{\parskip}{0pt}
  \setlength{\parsep}{0pt}
  \item \textbf{What, precisely, is being assured?} Specify the proposed
    rules and circumstances covered, and their justification. Distinguish
    proved guarantees from measured reliability and untested expectation.
  \item \textbf{What might the restriction cost us?} Assess potential harms
    from exclusions alongside harms from allowed outputs. Make uncertainties
    explicit; realize that refusals alone do not guarantee safety.
  \item \textbf{Who can question and repair the system?} Identify
    accountable people and institutions, and give them the information, time,
    and authority to challenge outcomes, revise policies, or suspend use.
    A human approval step alone---a token human-in-the-loop---does not constitute
    responsible oversight.
\end{enumerate}

We argue that promising general perfection is an ethical failure of precision;
invoking either that promise or its impossibility to shed responsibility is a
moral failure. Good guardrails support responsible judgment. They cannot
replace it. We must therefore retain the intellectual and moral
humility to recognize flaws in our machines, our rules, our policies, and even
ourselves---and the willingness to address those flaws.
\endgroup
\clearpage}{}{%
    \PackageError{manuscript}{Could not insert executive summary}{}%
  }
  \sslmaketitle
\else
  \maketitle
  \begin{abstract}
  \paperabstract
  \end{abstract}
  \clearpage
  
  \clearpage
\fi

\section{Introduction}


Any general assurance of a perfect guardrail is simply an impossibility.
Particular policies can be enforced exactly, but that does not support a
general promise of perfection. Nor does blocking more outputs, by itself,
settle the safety question: withholding permissible information may prevent
us from recognizing a danger or remedying a policy’s defects. We argue that
these limits demand precision in how we talk about the technology and both
individual and institutional responsibility for managing its risks. Neither
promises of perfection, nor the impossibility of fulfilling them, can
discharge that responsibility for any of us.

Language models are rarely deployed alone. Their outputs pass through
guardrails, which may include lists of banned phrases, classifiers trained to
recognize harmful content, a second model asked to judge whether an output is
acceptable, and, for models that call tools or run code, a monitor that
inspects each action and can stop the run. Guardrails make mistakes in both
directions. Forbidden outputs get through, often after a user rephrases a
prompt for that purpose, and permitted outputs get blocked. Anyone who builds,
buys, regulates, or relies on these systems needs to know whether those
mistakes are defects that better engineering will fix or whether, for some
policies, no guardrail can avoid them.

Guardrails are asked to prevent forbidden outputs while preserving what a
language model may legitimately provide. These are distinct demands. A guard
that blocks everything satisfies the first; a guard that passes everything
satisfies the second. The question is when a computable guard can satisfy
both, and what follows when it cannot. We call that combination exactness.
We use ``guard'' in the title in two senses: the computable guard of our formal
model, and the person or institution that stands guard over a deployed
system; we argue that neither, by its presence alone, discharges
responsibility for what a guardrail enforces.

The answer depends on the policy and on which outputs the generator can
produce. If every string is a possible output, an exact guard exists if and
only if the set of forbidden outputs is decidable
(Theorem~\ref{thm:general}). Under that assumption, whether a perfect
guardrail exists depends only on the policy, and no improvement in how the
guard is built can change the answer. A guard that is itself a language model
faces the same limit. A policy defined by a regular set of banned phrasings is
decidable. Idealized policies that judge an output by its truth, by what it
does when run, or by what it would lead an exact interpreter of instructions
to do are undecidable (Section~\ref{sec:policies}). For those policies, when
every string is a possible output, every guard gives the wrong verdict on
infinitely many outputs, and every guard that blocks all forbidden outputs
also blocks infinitely many permitted ones.

The results have practical consequences, which we develop in later sections. For
an undecidable policy, the trade-off between blocking too much and passing
too much cannot be engineered away. Whoever sets the policy has to choose
where that trade-off falls and should say so openly (Sections~5.1 and~5.2). A
requirement that a system never output~$X$, where $X$ is defined by meaning
or effect, cannot be met exactly by inspecting outputs. It is better written
as a tolerated error rate, under a stated distribution of use and a stated
adversary (Sections~\ref{sec:limits} and~5.2). Where exact enforcement
matters, it may be available elsewhere in the system. A monitor that observes
every action of a running program, and can stop each one, exactly enforces
any policy that names a non-empty set of harmful actions
(Sections~\ref{sec:monitor-cases} and~4.2). That favors controlling an
agent's actions at the boundary of the system over judging, in advance, the
text that requests them. Finally, a policy becomes a well-defined set only
once someone specifies it, and how it is specified determines whether it can
be enforced exactly (Section~5.1). None of this shows that guardrails are
ineffective. The results give no positive lower bound on average error, and
a guardrail can still be accurate enough to be useful in ordinary use
(Section~\ref{sec:limits}).

We bring guards, monitors, and several semantic policies into a
common account of this boundary. We also identify cases in which
exactness remains possible, including bounded outputs and fully observed,
interceptable actions. Those distinctions matter to policy: a guarantee of
compliance with a specified rule does not by itself establish the safety of
the wider activity, including the consequences of information withheld. Our
ethical argument concerns that further responsibility; we do not treat it
as an additional mathematical consequence.

We have a proof to offer, but as researchers offering modest contributions
to science, engineering, and policy, we must answer not just for the shape
of our proof but for how precisely we state what it warrants and implies.
We must consider what follows from the proof, what cannot follow from it,
and what \emph{should} not follow. Its conclusions can inform the design,
deployment, and governance of computational systems, and we have a
responsibility that they do so accurately.

\paragraph{Basis and related work.}
An earlier version of the mathematical argument appears in Long's
``No Perfect Guardrail''~\cite{long2026guardrail}.
The results are consequences of the undecidability of the halting
problem~\cite{turing1936}, G\"odel's first incompleteness
theorem~\cite{godel1931}, and Rice's theorem~\cite{rice1953}. We
state what these theorems imply for guardrails on the outputs of a language
model, under what assumptions, and what they do not imply.
Taken one at a time, several of the results are close to known ones; we
offer a characterization rather than a single undecidability result. For a
full generator, an exact guard exists exactly when the policy is decidable,
and if the policy is undecidable every sound guard excludes infinitely many
permitted outputs. We apply it to truth as well as to programs, judge
programs by the actions they perform rather than the results they return,
distinguish guards from monitors, and try to be precise about where
exactness remains possible.
Glukhov et al.~\cite{glukhov2023} apply Rice's theorem to the censorship of
outputs that describe programs and conclude that the set of permissible
outputs is undecidable, which gives the necessity half of
Theorem~\ref{thm:general}(a) for one class of policies. They also show that
censorship fails for outputs of bounded length when the recipient can invert
an encoding. This is consistent with Proposition~\ref{prop:finite} because
their policy depends on what the recipient can decode, while ours is a fixed
set of strings. Section~\ref{sec:persons} treats the recipient separately.
Their compositional Mosaic attacks, which combine several permitted outputs,
are outside our analysis.
Hamlen et al.~\cite{hamlen2006} show that a property of programs is
enforceable by static analysis if and only if it is decidable, and
Theorem~\ref{thm:general}(a) is the corresponding statement for guards that
see the prompt. Our treatment of monitors follows
Schneider~\cite{schneider2000}, Viswanathan~\cite{viswanathan2000}, who
required that the test a monitor applies be computable, and Hamlen et
al.~\cite{hamlen2006}. Monitors that may suppress or insert actions enforce a
larger class of policies~\cite{ligatti2005}. Cohen~\cite{cohen1987} showed
that no algorithm decides whether a program is a virus, and
Adleman~\cite{adleman1990} formalized the result. Alfonseca et
al.~\cite{alfonseca2021} reduce the halting problem to the question of whether
a machine will harm humans, the reduction behind
Corollaries~\ref{cor:harm} and~\ref{cor:recipient}. Melo et
al.~\cite{melo2024} apply Rice's theorem to the question of whether a model
is aligned and propose a judge that masks harmful outputs, and
Theorem~\ref{thm:general}(a) states when such a judge can be exact. Xu et
al.~\cite{xu2024} show that a computable model must err against some
computable ground truth, and Kalai and Vempala~\cite{kalai2024} give a
statistical lower bound on the rate of hallucination. Ball et
al.~\cite{ball2026} show, under cryptographic assumptions, that efficient
filters need not exist even where exact ones do, which bears on the question
of cost that Section~\ref{sec:limits} leaves open.
Theorem~\ref{thm:further}(b) is an instance of the result of de Leeuw et
al.~\cite{deleeuw1956} that machines with random choices and bounded error
decide only decidable sets.

Our ethical analysis also examines the safety consequences of conservative
exclusion (attempting to guarantee safety by being more cautious and excluding
more potential dangers). Hamlen et al.~\cite{hamlen2006} recognize that
conservative enforcement may reject policy-compliant programs, and Glukhov et
al.~\cite{glukhov2023} discuss restrictions that sacrifice usefulness to secure
enforceable constraints. We ask what follows when the excluded possibilities
may themselves be needed for safety. The censorship of permissible outputs
may unforeseeably withhold the means to recognize a danger, or to remedy a
policy's defects. The justification of a safety claim, and responsibility for
its realization, would therefore extend beyond mere issues of enforceability
to the adequacy of the policy and the consequences of any exclusions enforced
by its implementation.

Since no guard can enforce an undecidable policy exactly, such a policy
should be written on the assumption that whatever enforces it will make
errors. A well-written policy says which errors it tolerates, at what rate,
under what use and against which adversary, and who reviews the cases it
gets wrong. That recommendation is normative. The theorems show that the
errors exist, and they leave the choice of which errors to accept to the
people who write the policy.

\section{General Results and Their Scope}\label{sec:model}

We introduce the formal model and state the general results for
guards, which examine outputs before use, and monitors, which examine actions
during execution. We give the proof of the central guard theorem here;
supporting facts, further results, and proofs omitted or sketched in the body
appear in the \ref{app:proofs}. We then explain the scope and limitations
of these results. In Section~\ref{sec:policies}, we apply them to particular policies.

\subsection{Model}\label{sec:setup}

\paragraph{Notation.}
$\Sigma$ is a finite alphabet with at least two symbols, $\Str$ is the set of
finite strings over $\Sigma$, $\emptystr$ is the empty string, $|w|$ is the
length of the string $w$, and $\Sigma^{\le n}$ is the set of strings of length
at most $n$.  A \emph{language} is a subset of $\Str$.  The \emph{complement}
of a language $B$ is $\co B=\Str\setminus B$, and the \emph{symmetric
difference} of languages $B$ and $C$ is
$B\symdiff C=(B\setminus C)\cup(C\setminus B)$.  We write
$\N=\{0,1,2,\dots\}$.

We assume the definition of a Turing
machine~\cite{hopcroftullman1979}, and fix a standard coding under which every
string $e$ is the code of a machine $M_e$~\cite{rogers1967}.  We write
$M_e(x)\halts$ if $M_e$ halts on input $x$, and $M_e(x)\diverges$ if it does
not.  A function is \emph{computable} if some Turing machine halts on every
input $x$ with output $f(x)$; arguments and values that are not strings are
coded as strings.  A language $B$ is \emph{decidable} if the function that is
$1$ on $B$ and $0$ on $\co B$ is computable, and \emph{undecidable} otherwise.
It is \emph{recursively enumerable} if it is the set of inputs on which some
Turing machine halts.  A language $B$ is \emph{many-one reducible} to a
language $C$, written $B\mred C$, if there is a computable function $f$ such
that $x\in B$ if and only if $f(x)\in C$, for every $x$.  The \emph{halting
set} is $K=\{e:M_e(e)\halts\}$.

\begin{definition}[Generator, policy, guard]\label{def:guard}
A \emph{generator} is a computable function $G:\Str\times\Str\to\Str$.  The
string $G(u,r)$ is the output produced in response to the \emph{prompt} $u$
when the random choices of the generator are determined by the \emph{seed}
$r$.  The set of possible outputs on prompt $u$ is
$\Out_G(u)=\{G(u,r):r\in\Str\}$, and $\Out_G$ is the union of these sets over
all prompts.  The generator is \emph{full} if $\Out_G=\Str$.

A \emph{policy} is a language $B$, the set of \emph{forbidden} outputs.
Outputs in $\co B$ are \emph{permitted}.  A \emph{guard} is a computable
function $F:\Str\times\Str\to\{0,1\}$.  It \emph{blocks} the output $w$ on
prompt $u$ if $F(u,w)=1$ and \emph{passes} it if $F(u,w)=0$.  With respect to
$G$ and $B$, a guard is \emph{sound} if it blocks every forbidden output of
$G$ on every prompt, \emph{permissive} if it passes every permitted output of
$G$ on every prompt, and \emph{exact} if it is both.
\end{definition}

A guard is the formal counterpart of a guardrail, and an exact guard is a
perfect one.  The guard that blocks every output is sound, and the guard that
passes every output is permissive.  The question is whether one guard can be
both.  We take each policy as a given set; how a policy is specified is a
separate question.  A generator models a generation procedure that terminates
on every prompt and seed.  A guard may itself be a language model.

\subsection{Guards}\label{sec:guards}

\begin{theorem}\label{thm:general}
Let $G$ be a full generator and $B$ a policy.
\begin{enumerate}
\item[\textup{(a)}] An exact guard exists if and only if $B$ is decidable.
\item[\textup{(b)}] If $B$ is undecidable, every guard gives the wrong verdict
  on infinitely many outputs; every sound guard blocks infinitely many
  permitted outputs; and every permissive guard passes infinitely many
  forbidden outputs.
\end{enumerate}
\end{theorem}

\begin{proof}
Since $\Out_G=\Str$, the following search halts for every string $w$: list the
pairs $(u,r)$ in a fixed order and stop at the first with $G(u,r)=w$.  Let
$\pi(w)$ be the prompt of that pair.  Then $\pi$ is computable and
$w\in\Out_G(\pi(w))$.  For a guard $F$ let $D_F=\{w:F(\pi(w),w)=0\}$, which is
decidable.  If $F$ is sound then $D_F\subseteq\co B$, and if $F$ is permissive
then $\co B\subseteq D_F$.

(a)~If $B$ is decidable, the guard that blocks $w$ if and only if $w\in B$ is
exact.  If $F$ is exact then $D_F=\co B$, so $B$ is decidable.

(b)~Let $E=D_F\symdiff\co B$, so that $\co B=D_F\symdiff E$.  If $E$ were
finite it would be decidable, and then $\co B$ would be decidable.  So $E$ is
infinite.  On the prompt $\pi(w)$, each $w\in E$ is either forbidden and
passed, or permitted and blocked.  If $F$ is sound then $E=\co B\setminus D_F$,
and if $F$ is permissive then $E=D_F\cap B$.

\end{proof}

The theorem has the following limits.  To conclude that no exact guard exists,
the generator need not be full: it suffices that $K$ is many-one reducible to
$B$ or to $\co B$ by a function whose every value the generator can emit.
Part~(b) counts strings and does not bound an average error
(Section~\ref{sec:limits}).

Three further consequences are stated and proved as Theorem~\ref{thm:further}
in Section~\ref{app:more} of the Appendix: if $B$ is infinite, no algorithm decides from the
code of a guard whether the guard is sound; if $B$ is undecidable, every
randomised guard gives the wrong verdict with probability at least
$\tfrac12$ on some output; and if $\co B$ is not recursively enumerable, every
generator that emits only permitted strings fails to emit infinitely many of
them.

\subsection{Monitors}\label{sec:monitors}

A guard examines a program before it is run.  A monitor examines the actions
of a program while it runs, and can stop the program.

An output that is run as a program has effects while it runs, and not only a
final result.  Fix a finite non-empty alphabet $A$ of \emph{actions}, such as
commands to a device, calls to an operating system, and symbols written to a
file.  Let $A^{*}$ be the set of finite sequences of actions, $A^{\omega}$ the
set of infinite sequences, and $\Traces=A^{*}\cup A^{\omega}$.

\begin{definition}[Trace policy, monitor]\label{def:monitor}
A \emph{trace policy} is a set $\Bad\subseteq A^{*}$ that does not contain the
empty sequence and that contains $sz$ whenever it contains $s$.  A finite
trace is \emph{forbidden} if it is in $\Bad$ and \emph{permitted} otherwise.
A \emph{monitor} is a computable function $\mu:A^{*}\to\{0,1\}$.  When a
program run under the monitor has performed the sequence $s$ and is about to
perform the action $a$, the program is stopped if $\mu(sa)=1$, and $a$ is
performed if $\mu(sa)=0$.  A monitor is \emph{sound} if $\mu(sa)=1$ whenever
$s$ is permitted and $sa$ is forbidden, \emph{permissive} if $\mu(s)=0$
whenever $s$ is permitted, and \emph{exact} if it is both.
\end{definition}

\begin{theorem}\label{thm:monitor}
An exact monitor exists if and only if $\Bad$ is decidable.  If $\Bad$ is
undecidable, then under every sound monitor there are infinitely many
permitted finite traces that cannot be performed in full, and under every
permissive monitor there are infinitely many forbidden finite traces that can.
\end{theorem}

The proof, in the \ref{app:proofs}, follows that of
Theorem~\ref{thm:general}: the finite traces that a monitor allows to be
performed in full form a decidable set, and if the monitor is exact this set
is the complement of $\Bad$.

A trace policy corresponds to a safety property in the sense of
Schneider~\cite{schneider2000}, who showed that a monitor that can only stop a
program enforces no other kind of policy.  The condition that $\Bad$ be
decidable corresponds to the condition that Hamlen et
al.~\cite{hamlen2006} place on the test that a monitor applies to finite
sequences of events.

\subsection{Scope of the general result}\label{sec:limits}

\paragraph{Bounded outputs.}\label{sec:finite}
A language model that stops at a limit of length is a generator, and it is not
full.  If the outputs of a generator are in $\Sigma^{\le n}$, the set of its
forbidden outputs is finite, so it is decidable, and an exact guard exists.
Theorem~\ref{thm:general} does not apply.  For an undecidable policy
$B$, each language $B\cap\Sigma^{\le n}$ is decidable, but no algorithm
produces from $n$ a machine that decides it
(Proposition~\ref{prop:finite} in Section~\ref{app:more} of the Appendix).

The proposition does not say that a particular bounded system lacks an exact
guard, or that its exact guard is costly.  Fullness is an additional
assumption about which outputs the generator can produce; allowing unlimited
memory does not establish it.

\paragraph{Average error.}
Theorem~\ref{thm:general}(b) counts strings.  It gives no positive lower bound
on the average error of a guard under a probability distribution on outputs.
Theorem~\ref{thm:further}(b) is a different statement: it bounds the error of
a randomised guard
on one particular output.  Let $P$ be a probability distribution on $\Str$ and
let
$\delta>0$.  There is a finite set $U$ of strings with $P(U)>1-\delta$.  The
guard that passes the permitted members of $U$ and blocks every other string
is sound, and the probability that it blocks a permitted output drawn from $P$
is less than $\delta$.  This holds for every policy.  If the policy is
undecidable, no single algorithm produces such a guard from every finite set
$U$, since applied to $U=\{w\}$ it would determine whether $w$ is permitted.

\paragraph{Adversaries.}
A small average error under $P$ does not bound the loss when outputs
are chosen by an adversary.  Suppose that a use of the generator is made by an
adversary with probability $\alpha$; that the adversary then obtains, with
probability $\sigma$, a forbidden output that the guard passes; that the loss
from such an output is at least $\lambda$; and that no loss is negative.  Then
the expected loss in one use is at least $\alpha\,\sigma\,\lambda$, whatever
the average error under $P$.  These quantities are assumed and are not
derived from Definition~\ref{def:guard}, and the bound holds for decidable
policies also.  If the guard is sound then $\sigma=0$; that $\sigma=0$ for a
particular adversary does not imply that the guard is sound.

\section{Special cases}\label{sec:policies}

One might hope that the general result is loose in practice: that the
policies we actually care about are decidable, so that perfect guards for them
exist.  We test that hope on policies of increasing depth: what an
output says (its form, Section~\ref{sec:lexical}), whether it is true
(Section~\ref{sec:truth}), what it does when run as a program
(Section~\ref{sec:programs}), and what it leads a recipient to do
(Section~\ref{sec:persons}).  The unrestricted semantic policies considered
below are undecidable; the policy on form is decidable. Restricted semantic
cases can also be decidable, as the qualifications below explain.
In Section~\ref{sec:monitor-cases}, we ask the same question of monitors,
distinguishing interception of a specified action from classification of a
delivered string under an undecidable policy. We summarise the results in
Table~\ref{tab:summary}.

\subsection{Form: banned phrasings}\label{sec:lexical}

Let $L$ be a regular language of banned phrasings, and let $B_L$ be the set of
strings in which some member of $L$ occurs as a contiguous part.  Then $B_L$
is regular, and a finite automaton decides whether $w\in B_L$ in $|w|$
steps~\cite{hopcroftullman1979}.  By Theorem~\ref{thm:general}(a) an exact
guard exists.  A finite list of banned phrases is a special case.

Regular languages are Type~3 of the hierarchy of
Chomsky~\cite{chomsky1959}.  A list of banned words takes no account of
context.  A regular language can take some: the set of strings in which a
given word occurs without a second given word immediately after it is regular.
Other dependence on context is not regular.  The set of strings over an
opening and a closing bracket in which the brackets are correctly nested is
context-free (Type~2) and not regular.  Every context-free language is
decidable~\cite{hopcroftullman1979}.

\subsection{Truth: false sentences of arithmetic}\label{sec:truth}

A \emph{sentence of arithmetic} is a sentence of the first-order language with
the symbols $0$, $S$ (successor), $+$, $\times$, and $<$.  It is \emph{true}
if it holds in $\N$.  Fix a standard effective G\"odel coding of formulas by
strings~\cite{godel1931}.  Let $\Sent$ be the set of strings that code
sentences, and $\Th\subseteq\Sent$ the set of strings that code true
sentences.  $\Sent$ is decidable, and the code of the negation of a sentence
is computable from the code of the sentence.  A policy $B$ is a
\emph{falsehood policy} if $B\cap\Sent=\Sent\setminus\Th$, that is, if among
the sentences of arithmetic it forbids every false one and no true one.

\begin{theorem}\label{thm:truth}
No falsehood policy is decidable, and none is recursively enumerable or has a
recursively enumerable complement.
\end{theorem}

The proof is given in the \ref{app:proofs}.  By G\"odel's first
incompleteness theorem~\cite{godel1931}, no machine lists the true sentences
of arithmetic, and a machine that listed either the forbidden or the
permitted strings of a falsehood policy would yield such a machine.

The theorem concerns a policy that classifies every sentence of arithmetic
correctly.  It does not say that every policy about false statements is
undecidable: truth in the first-order arithmetic of addition alone is
decidable~\cite{presburger1929}.  For Theorem~\ref{thm:general} to apply it
suffices that the generator can emit the code of every sentence.  By
Theorem~\ref{thm:further}(c), a generator that emits only true sentences of
arithmetic fails to emit infinitely many true sentences.

\subsection{Behavior: programs}\label{sec:programs}

\begin{definition}[Action machine, trace, behavior]\label{def:action}
An \emph{action machine} is a deterministic Turing machine with an input tape,
work tapes, and a write-only action tape over $A$ on which the head moves only
to the right.  To write a symbol on the action tape is to perform that action.
Fix a standard coding under which every string $p$ is the code of an action
machine $N_p$.  We call $p$ a \emph{program}.  The \emph{trace} $\tr_p(x)$ is
the finite or infinite sequence of actions that $N_p$ performs on input $x$,
in order.  It is defined whether or not $N_p$ halts.  The \emph{behavior} of
$p$ is the function $\beta_p:\Str\to\Traces$ that maps $x$ to $\tr_p(x)$.  The
\emph{silent behavior} is the behavior that performs no action on any input.
A policy $B$ is \emph{behavioral} if $\beta_p=\beta_{p'}$ implies that $p$
and $p'$ are both in $B$ or both outside it.
\end{definition}

This is a simplified trace-based model of programs, following Hamlen et
al.~\cite{hamlen2006}.  A behavioral policy judges a program by the actions
it performs and by nothing else.  A policy that forbids programs that perform
a given action within a given number of steps is not behavioral, and it is
decidable by running the program for that number of steps.

\begin{theorem}[Rice's theorem for traces]\label{thm:rice-traces}
Let $B$ be a behavioral policy that is neither empty nor all of $\Str$.  If
no program with the silent behavior is in $B$, then $K\mred B$.  Otherwise
$K\mred\co B$.  In both cases $B$ is undecidable.
\end{theorem}

Rice's theorem~\cite{rice1953} is the same statement for Turing machines that
are judged by the result they return on each input on which they halt.  As
stated it does not mention actions, and a program can perform a harmful action
and never return a result.  We therefore state the theorem for traces.  The
proof, given in the \ref{app:proofs}, is that of Rice's theorem.  From
$e$ one computes a program that first simulates $M_e$ on input $e$ and then,
if the simulation halts, behaves like a fixed program $q\in B$.  It has the
behavior of $q$ if $e\in K$ and the silent behavior otherwise.  The argument
applies to traces because the program performs no action while it waits.

\begin{corollary}\label{cor:harm}
Let $H\subseteq A$ be a non-empty set of harmful actions, let $x_0\in\Str$,
and let $\Harm_H$ be the set of programs $p$ such that an action in $H$ occurs
in $\tr_p(x_0)$.  Then $\Harm_H$ is recursively enumerable, $K\mred\Harm_H$,
$\Harm_H$ is undecidable, and its complement, the set of safe programs, is not
recursively enumerable.  The same holds for the set of programs that perform
an action in $H$ on some input.
\end{corollary}

\begin{proof}
$\Harm_H$ is behavioral.  It contains no program with the silent behavior,
and it contains a program that performs an action in $H$ on every input.
Apply Theorem~\ref{thm:rice-traces} and fact (F3) of
Section~\ref{app:facts} of the Appendix.  A machine that runs $N_p$ on
$x_0$ and halts when an action in $H$ is performed halts exactly on the
members of $\Harm_H$.  For the last statement, the machine runs $N_p$ on all
inputs in turn, for increasing numbers of steps.
\end{proof}

\paragraph{Restricted programs.}
Suitable restrictions on the programs that a generator emits can restore
decidability.  If every program emitted halts within a number of steps
computable from its text, membership in $\Harm_H$ is decidable for those
programs by running them.  A restriction on the format of programs that does
not limit what they compute does not have this effect.

\subsection{Effect on a recipient: an idealized recipient}\label{sec:persons}

An output may do harm through whoever receives it.  A \emph{recipient} is a
function $\rho:\Str\to\Traces$, where $\rho(w)$ is the sequence of actions
that the recipient performs upon receiving the message $w$.  The recipient is
not assumed to be the behavior of any program.  For a non-empty set
$H\subseteq A$ of harmful actions, the policy \emph{induced} by $\rho$ is the
set $B_H(\rho)$ of messages $w$ such that an action in $H$ occurs in
$\rho(w)$.  A recipient is \emph{instructable} if there is a computable
function $\iota$ such that $\rho(\iota(p))=\tr_p(\emptystr)$ for every program
$p$.  The message $\iota(p)$ instructs the recipient to carry out the program
$p$, and the recipient does so exactly.

\begin{corollary}\label{cor:recipient}
If $\rho$ is instructable, then $K\mred B_H(\rho)$.  Hence $B_H(\rho)$ is
undecidable and the set of harmless messages is not recursively enumerable.
If in addition $\rho$ is the behavior $\beta_c$ of some program $c$, then
$B_H(\rho)$ is recursively enumerable.
\end{corollary}

The proof, in the \ref{app:proofs}, composes the encoding $\iota$ with
the reduction in the proof of Theorem~\ref{thm:rice-traces}: the message that
instructs the recipient to carry out a program leads to a harmful action
exactly when that program performs one on the empty input.

An instructable recipient is an interpreter of programs that is exact and
unlimited in time and memory.  The corollary applies to a person only if there
is an encoding $\iota$ of instructions that the person carries out in this
way.  It is not a theorem about persuasion.

\subsection{Monitoring a running program}\label{sec:monitor-cases}

For a monitor, the condition of Theorem~\ref{thm:monitor} is that $\Bad$ be
decidable.  Whether it holds depends on what the monitor observes
and on what counts as the violation.  There are two cases.

\paragraph{The violation is an action.}
Let $H\subseteq A$ be a non-empty set of harmful actions, and let $\Bad$ be
the set of finite traces in which an action in $H$ occurs.  This set is
decidable.  The monitor that stops a program exactly when its next action is
in $H$ is exact.  By Corollary~\ref{cor:harm} no guard decides from the text
of a program whether the program will perform an action in $H$.  The monitor
prevents every such action and stops no program whose trace is permitted.
This requires that the harmful actions be specified as a set of symbols, and
that the monitor observe every action and be able to stop it.

\paragraph{The violation is the delivery of a string.}
Suppose that $\Sigma\subseteq A$ and that $A$ contains a symbol $\dlv$ that is
not in $\Sigma$.  The action $\dlv$ \emph{delivers} the string of symbols of
$\Sigma$ written since the last action that was not in $\Sigma$, or since the
start if there was none: it hands that string to a recipient who reads it, or
to a machine that runs it.  Let $B$ be a policy on delivered strings, and let
$\Bad_B$ be the set of finite traces in which a string in $B$ has been
delivered.

\begin{proposition}\label{prop:delivery}
$B\mred\Bad_B$.  Hence if $B$ is undecidable, no monitor is exact with respect
to $\Bad_B$.
\end{proposition}

\begin{proof}
For every string $w$, the trace $w\dlv$ is in $\Bad_B$ if and only if
$w\in B$.
\end{proof}

In this case the monitor must decide, before a string is delivered, whether
the string is in $B$.  Some harmful programs perform only operations that are
individually ordinary, such as writing a file, and the harm is in what is
written.  If the violation is not the delivery but a later action of whoever
receives the string, and that action is observed by a monitor, the first case
applies to that action.

\subsection{Summary}\label{sec:summary}

Table~\ref{tab:summary} collects the results.  If every string is a possible output, an exact guard is a decision procedure
for the set of forbidden outputs.  There is none for a policy that forbids
every false sentence of arithmetic and no true one; for a behavioral policy
on programs that forbids some programs and permits others; or for a policy
that forbids the messages that lead an instructable recipient to a harmful
action.  The same holds for a monitor that must decide, before a string is
delivered, whether the string is forbidden by such a policy.  The results do
not exclude exact guards for outputs of bounded length or for suitably
restricted programs, or exact monitors for policies on observed actions, and
they do not bound average error under an arbitrary probability distribution on
outputs.

\begin{table}[!htbp]
\centering
\small
\renewcommand{\arraystretch}{1.15}
\caption{Summary of the results.  The general results reduce the existence of
an exact guard or monitor to the decidability of the policy that it enforces;
the specific cases place particular policies on either side of that boundary.}
\label{tab:summary}
\begin{tabularx}{\textwidth}{@{}>{\raggedright\arraybackslash}X>{\raggedright\arraybackslash}p{3.4cm}>{\raggedright\arraybackslash}X>{\raggedright\arraybackslash}p{1.5cm}@{}}
\toprule
Mechanism and policy & Decidable? & Consequence & Result\\
\midrule
\multicolumn{4}{@{}l}{\emph{General results}}\\
Guard on a full generator; any policy $B$
  & Depends on $B$
  & An exact guard exists if and only if $B$ is decidable; otherwise every
    guard gives the wrong verdict on infinitely many outputs
  & Thm.~\ref{thm:general}\\
Monitor; any trace policy $\Bad$
  & Depends on $\Bad$
  & An exact monitor exists if and only if $\Bad$ is decidable
  & Thm.~\ref{thm:monitor}\\
\addlinespace
\multicolumn{4}{@{}l}{\emph{Specific policies}}\\
Banned phrasings (a regular language)
  & Yes
  & An exact guard exists
  & \S\ref{sec:lexical}\\
False sentences of arithmetic
  & No; neither the policy nor its complement is r.e.\ (G\"odel)
  & No exact guard
  & Thm.~\ref{thm:truth}\\
Behavioral policy on programs, neither empty nor all of $\Str$
  & No (Rice's theorem for traces)
  & No exact guard
  & Thm.~\ref{thm:rice-traces}\\
Programs that perform a harmful action
  & No; the harmful programs are r.e., the safe ones are not
  & No exact guard
  & Cor.~\ref{cor:harm}\\
Messages that lead an instructable recipient to a harmful action
  & No
  & No exact guard
  & Cor.~\ref{cor:recipient}\\
Monitor; the violation is an observed action
  & Yes
  & An exact monitor exists
  & \S\ref{sec:monitor-cases}\\
Monitor; the violation is the delivery of a string in an undecidable $B$
  & No
  & No exact monitor
  & Prop.~\ref{prop:delivery}\\
\addlinespace
\multicolumn{4}{@{}l}{\emph{Where exactness is not excluded}}\\
Outputs of bounded length
  & Yes for each bound, but no algorithm produces the decider from the bound
  & An exact guard exists for each bound
  & Prop.~\ref{prop:finite}\\
Programs that halt within a number of steps computable from their text
  & Yes
  & Harmfulness is decided by running them
  & \S\ref{sec:programs}\\
Average error under a distribution $P$ on outputs
  & Not required
  & For every $\delta>0$, some sound guard blocks a permitted output with
    probability less than $\delta$; for an undecidable policy, no single
    algorithm produces such guards from every finite set of outputs
  & \S\ref{sec:limits}\\
\bottomrule
\end{tabularx}
\end{table}

\FloatBarrier

\section{Interpreting the Guarantees}\label{sec:interpretation}

Guardrails are asked to do more than act as a ``nice to have'' watchdog for
a language model. We now consider a technical interpretation of guardrails
and what they can be asked to do, and how we should view the guarantees they can provide.

\subsection{Control and content}

Guardrails are asked to preserve the authority of intended
instructions when the material being processed can influence the processing
itself. Prompt injection makes this difficulty visible: content supplied for
a task can be treated as instructions governing the task. The National Cyber
Security Centre distinguishes this problem from injection attacks whose
instruction--data boundary can be enforced by the execution
mechanism~\cite{chismon2025prompt}.

That responsiveness is also a feature of the tool. We ask language models to
interpret unfamiliar instructions and arguments, and to develop approaches
not specified beforehand. Restricting what may influence their operation can
restrict what they help us discover. The comparison with self-modifying code
concerns changes in effective instructions and subsequent behavior; it does
not require inference to rewrite model weights or executable code.
The security literature offers a related caution: return-oriented programming
and ``weird machines'' show that a system can be driven to compute through
behaviors outside its intended interpretation~\cite{shacham2007,dullien2020}.
An output need not be written as a program to be run as one, so policies that
judge outputs by what they do when run (Section~\ref{sec:programs}) can apply
to outputs that were never intended as code.

These observations explain the relevance of semantic output policies. They
do not establish the assumptions of an undecidability reduction. In
particular, an agent that reads documents is not thereby the instructable
recipient of Corollary~\ref{cor:recipient}: that idealisation faithfully
executes every encoded program without limits on time or memory. Susceptibility
to prompt injection or human persuasion is a different claim. Empirical work
on sycophancy and conversational persuasion gives separate reasons to examine
how language models interact with human judgment~\cite{sharma2024,salvi2025}; it does
not establish the recipient idealisation.

\subsection{Where guarantees remain possible}

Theorem~\ref{thm:general} rules out exact discrimination for an undecidable
policy on a full generator. It does not rule out sound guards, useful
restrictions, or proofs of particular safety properties. A sound guard may
block every output. A less restrictive one may admit a verified subset. Under
the theorem's assumptions, however, every sound guard excludes infinitely
many permitted outputs. Whether those exclusions are acceptable is not
settled by soundness.

This boundary is consistent with formal verification. A proof of a
particular system, such as a verified operating-system kernel~\cite{klein2009},
is not a procedure that decides the correctness of every unrestricted
program. Finite-state model checking~\cite{clarke1986}, supplied proofs, and
restrictions on the programs or properties under consideration can establish
specific guarantees. Abstract interpretation can instead provide sound
approximations that admit false alarms~\cite{cousot1977}. Automating code
production does not remove the need to establish what such verification
covers~\cite{symonshorner2020}.

The same care is needed when relating the model to deployment. A fixed
output-length bound admits an exact guard in principle. For an undecidable
policy, no algorithm constructs these exact guards uniformly from the bound,
but this does not show that a particular bounded guard is unavailable or
costly. Results on saturated and log-precision transformers demonstrate how
computational power depends on architectural and arithmetic
assumptions~\cite{merrill2022,merrillsabharwal2023}. Idealized models with unbounded context or external memory can support
universal computation~\cite{perezbarcelo2021,schuurmans2023}; this alone does not
establish that a generator can emit every string, or the family of strings
needed for a particular reduction. Those reachability assumptions must be
justified for the claim being made.

Monitoring also changes the question. In Section~\ref{sec:monitor-cases}, we give an
exact monitor for specified actions that are fully observed and interceptable.
When the violation instead consists in delivering a string forbidden by an
undecidable policy, the classification problem remains. Confinement and
side-channel research~\cite{lampson1973,kocher2019} gives reasons to examine
whether the observation and interception assumptions hold; it does not
establish that every possible confinement mechanism must fail.

\subsection{What evidence must establish}

The infinite-error result supplies no positive lower bound on average error
under an arbitrary output distribution, nor does it determine the likelihood
or cost of a successful attack. Measurements of those quantities can therefore
matter greatly to deployment decisions without refuting the theorem.
Likewise, probes and representation-level interventions can provide evidence
about particular systems~\cite{markstegmark2023,zou2023}. Work on the empirical
and conceptual limitations of lie detectors cautions against treating such
evidence as a general guarantee~\cite{levinsteinherrmann2024}. A claim that a
detector covers unanticipated forms of deception needs support beyond its
observed successes.
Rice's theorem cannot supply that conclusion merely from the word
``deception'': the relevant property and class of computations must first be
specified.

The distinction is between establishing a guarantee under stated assumptions
and justifying reliance on that guarantee in an application. In the next section,
we address that further responsibility, including the consequences of what
a safeguard prevents.

\section{Ethical and Policy Implications}\label{sec:amer-suggestions}
\begingroup
\clubpenalty=10000
\widowpenalty=10000


Our responsibility in presenting an impossibility result includes resisting
the temptation to make it settle more than it does. ``No exact guard for this
policy and output domain'' is a mathematical conclusion. ``This deployment is
safe enough,'' ``this restriction is justified,'' and ``this technology must
be prohibited'' are judgments requiring further evidence and argument. The
proof does not decide them for us.

The ethical obligation is not peculiar to AI. Work on the ethics of
belief~\cite{clifford1877ethics,feldman1988obligations,feldman2000ethics} asks how beliefs should be
formed, maintained, and acted upon. The modest proposition needed here is
that \emph{how one inquires can itself be normatively assessable}. Arriving at
a defensible conclusion does not automatically vindicate the process by which
one arrived there. Researchers who offer a result for use in engineering and
policy must make its assumptions and limits available to those who will rely
on it. Those who invoke the result bear a corresponding responsibility to
understand what it warrants. In software-intensive science, the grounds for
reliance on computational results are themselves a subject of
inquiry~\cite{symonshorner2014}.

That obligation applies to advocates and critics alike. Evidence on motivated
reasoning, confirmation bias, and selective
exposure~\cite{kunda1990case,nickerson1998confirmation,hart2009validated} gives
reason to scrutinise how congenial conclusions are reached; it does not
establish bad faith in any particular actor. Studies of biased assimilation
and motivated skepticism show how prior commitments can influence the
evaluation of evidence~\cite{lord1979biased,taber2006motivated}. An impossibility theorem can be
recruited to justify a prohibition it does not entail, or dismissed because
its consequences are inconvenient. Neither the prospect of benefit nor the
fear of harm excuses an unsupported inference. Precision matters because
other people may act on the assurance or warning we give them.

\subsection{Policy compliance and the consequences of exclusion}

The formal model takes a policy as a given set of forbidden outputs. Whether
that set adequately captures what ought to be prevented is a further
question. AI safety research already recognizes failures of objective
specification, including reward hacking and specification
gaming~\cite{amodei2016concrete,krakovna2020specification}. Guaranteed policy
compliance is not guaranteed overall safety. A
policy can fail to capture what safety requires, and its enforcement can
suppress information needed to recognize or respond to danger. For a
general-purpose informational tool,
these exclusions cannot simply be treated as a loss of usefulness with no
bearing on safety. Filtering can suppress the very novelty the tool is
intended to enable.

Under the assumptions of Theorem~\ref{thm:general}, every sound guard for an
undecidable policy excludes infinitely many permitted outputs. The theorem
counts strings; it does not establish that those strings contain valuable
discoveries or safety-critical information. But neither does compliance
establish the harmlessness of the exclusions. A research tool might withhold
a legitimate hypothesis needed to find a remedy, or a security tool might
exclude an unfamiliar approach needed to recognize a vulnerability. These
possibilities illustrate why omission belongs in the safety assessment; they
are not findings that a particular filter has withheld such information.

It might be argued that, where the stakes are sufficiently high, any
possibility of an unsafe outcome would be intolerable. But that concern must
also extend to the unforeseeable censorship of permissible outputs:
withholding a warning, remedy, or unfamiliar approach may itself be a source
of unanticipated danger that the policy does not cover. A demand for safety
cannot count the dangers of disclosure while presuming the harmlessness of
exclusion. The higher the stakes, the greater our responsibility to examine
both. This does not establish that every exclusion is dangerous, or that
unrestricted release is safe. It establishes that enforcement of guardrails,
no matter how conservative, cannot alone settle the safety question.

Disclosing that filtering occurs does not disclose what has been lost. A user
may consent to a restriction without being able to inspect the possibilities
it excludes. Such consent does not by itself establish that inaccessible
possibilities were expendable, or that a blind spot created in the name of
safety is itself safe. Nor does a large volume of useful accepted outputs
settle the consequences of what remains unavailable.

The relation to the proof can be stated conditionally. Suppose a proposed
guarantee requires a sound guard also to preserve every permitted output whose
withholding could cause harm. If one guard must satisfy that requirement
regardless of \emph{which} permitted output is indispensable, it must pass
every permitted output and is therefore exact. Theorem~\ref{thm:general} rules
this out for a full generator and an undecidable policy.

This demanding requirement is not necessary for every narrower guarantee.
Several outputs may serve the same need, so a guarantee might require only
that \emph{some} output serving each relevant need be passed. A general
argument may also cover cases that were not individually foreseen. Such
guarantees need not demand flawless prescience, and the theorem does not
establish that these weaker requirements are impossible to meet. But the
scope of the assurance still needs justification: what needs and circumstances
does it cover, and what grounds do we have for treating the exclusions as
compatible with safety within that scope?

Conservative exclusion cannot automatically be counted as the harmless side
of error. A claim that compliance suffices for safety must explain why the
policy and its enforcement are adequate to the circumstances for which that
claim is made, including the consequences of withholding information.
Presenting an assurance as more comprehensive than its grounds support is an
ethical failure of precision; invoking it to abdicate responsibility is a
moral failure.

\subsection{Making responsibility practicable}

The distinction between policymaking and rule enforcement is therefore
essential. The policymaker must justify what should be pursued or protected;
the rule maker and enforcer translate that purpose into constraints and apply
them. Successful enforcement cannot validate the purpose it serves. Likewise,
the researcher must establish what a finding warrants, while the user must
judge whether a particular application is justified. A research result does
not authorise every use, and user approval does not validate the research.
These are roles, not fixed identities. Distributing the work must not erase
the responsibilities attached to it. The analysis of agency
laundering~\cite{rubel2019agency} identifies a related wrong: using a
technological intermediary to obscure responsibility and obstruct demands for
an account.

Keeping a human at the controls is not sufficient. Bainbridge's account of
the ironies of automation already showed how automation can leave operators
with difficult demands that its design inadequately supports~\cite{bainbridge1983ironies}.
A person can approve each
operation while accepting purposes and conclusions they have ceased to
examine. Conversely, substantial delegation can be compatible with responsible
reliance when its grounds, scope, and conditions for reconsideration are
understood. Research on cognitive offloading and automation
bias~\cite{risko2016offloading,parasuraman2010complacency} helps explain why
participation alone is an inadequate assurance. Lee et al.'s survey of knowledge
workers found an association between confidence in generative AI and reduced
self-reported critical-thinking effort~\cite{lee2025critical}; this is evidence
about reported practice, not a demonstration of inevitable cognitive decline.
The relevant obligation is to
exercise judgment appropriate to one's role and capacities, not to reproduce
every computation or predict every result. Failure to exercise that judgment
can itself be culpable; it does not make responsibility disappear.

Institutions must make this responsibility practicable. An overseer needs
enough information, time, and authority to question, refuse, or revise the
activity. Assigning a person blame after a failure does not establish that
these conditions existed. Developers and deployers remain answerable for the
arrangements within which others are asked to rely on their systems.

Elish's ``moral crumple zones'' and Green's critique of human-oversight policies
make this institutional problem explicit~\cite{elish2019moral,green2022flaws}.
Accounts of meaningful human control connect responsibility to responsiveness
to relevant reasons and to the capabilities of the people in the
system~\cite{santonidesio2018meaningful,cavalcantesiebert2023meaningful}.
Experiments on cognitive forcing also show that interface design can reduce
overreliance, with costs to users' experience and unequal benefits across
participants~\cite{bucinca2021trust}. Responsible oversight therefore requires
conditions that can be examined and improved.

A defensible assurance should identify the policy being enforced, the output
domain or actions covered, the assumptions about observation and
interception, and the evidence for performance beyond a formal guarantee's
scope. It should distinguish a proved property from measured reliability,
describe known exclusions and uncertainties about their consequences, and say
who can respond when those assumptions or consequences are challenged.
``Usually works'' is an empirical claim to assess, not a substitute for
specifying what works and under what conditions.

Policy then has to address the whole application: the harms of use, the harms
of restriction or non-use, and the means of detecting and correcting mistakes.
This is consistent with a systems approach to safety that examines interactions
and organisational arrangements rather than locating every failure in an
isolated component~\cite{leveson2004accident}. Work on language-model risks and
release strategies similarly places deployment in a wider social
setting~\cite{weidinger2021,solaiman2019release}.
In practice, this calls for monitoring after deployment, routes for contesting both accepted
and refused outputs, and authority to revise or suspend an arrangement when
its justification no longer holds. Reviewing exclusions does not require
indiscriminate disclosure of dangerous material; it requires ways to examine
their consequences without presuming them harmless. The question is: safe from
what, for whom, under what circumstances, by what mechanism, with what
remaining risk, and who is responsible for responding?

\subsection{Humility and revisability}

Even exact enforcement of a particular rule leaves someone responsible for
choosing it, judging its adequacy, examining its exclusions, and revising it
as circumstances change. This obligation survives the cases in which the
formal results permit exact guards or monitors. ``Guardrails never being
enough'' is, in this sense, a natural condition to expect and accept.
Improving controls remains worthwhile; promising that they can eliminate the
need for responsible judgment mistakes their function.

Wiener's warning about the monkey's paw connects literal fulfillment of a
wish to the consequences its maker failed to consider. He extends the warning
to human institutions, where people can be used as components of a machine
rather than treated as responsible human beings~\cite[pp.~212--213]{wiener1950human}.
The difficulty concerns both the means of carrying out a purpose and our
judgment about the purpose itself. We may express what we want imprecisely;
we may also come to recognize a flaw in what we sincerely wanted. More
faithful execution cannot settle whether that purpose deserves to be pursued.

This matters to the ability to correct policy. A restriction might suppress
the warning or unfamiliar argument that would help expose a defect in the
restriction itself. The absence of criticism among the outputs it permits
would then be poor evidence of its adequacy. This is a possible institutional
failure, not a consequence established by the theorem. It gives a further
reason to preserve routes of challenge that do not depend solely on what the
existing policy allows people to see. Work on epistemic injustice and
resistance draws attention to whose testimony is credited and whose
interpretations can enter shared inquiry~\cite{fricker2007epistemic,medina2013epistemology}.
Providing a formal route for objection is insufficient if those who recognize
a defect cannot make themselves heard.

Revisability requires neither accepting every objection nor assuming every
change is an improvement. It requires that no policy acquire immunity from
reconsideration merely because it has been precisely expressed or
successfully enforced. A narrower guarantee can be valuable without settling
the worth of the whole arrangement. Decisions to deploy, restrict, or refrain
must remain answerable to evidence and to those who bear their consequences.
Intellectual ability alone does not provide immunity from myside
bias~\cite{stanovich2013myside}.

\section{Concluding Remarks}

Our formal results establish when exact enforcement is possible within the
models we study, and when it is not. Our broader ethical and policy arguments
concern the responsibility that remains in either case.

The impossibility of a general assurance of perfect enforcement is itself a
reason that responsibility cannot be discharged by invoking guardrails.
Promising such perfection is an ethical failure of precision; invoking either
that promise or the impossibility of fulfilling it to abdicate responsibility
is a moral failure. Responsible AI policy demands neither doomsaying nor
demands for the impossible. It demands precision in our discourse, and both
individual and institutional responsibility for managing risks that cannot be
wished away. Even exact enforcement cannot, by itself, assure the realization
of a policy’s purposes or establish their adequacy. Continued critical
challenge is both an essential element of good policy and a moral imperative.

We must cultivate and maintain the intellectual and moral humility to accept
that flaws can exist in our machines, our rules, our policies, or even in
our very selves. When we lose that humility and the willingness to repair any
part of our overall system, we invite a humbling lesson in the costs of our
own blindness. That obligation for precision and humility applies to the
claims we make in this paper as well as to the systems we examine. Good
guards can build and maintain good guardrails that support and encourage the
exercise of responsibility. But their mere presence can neither establish
the safety of our overall systems, nor dismiss each individual's agency and
responsibility to seek and maintain that safety.
\endgroup

\bibliographystyle{IEEEtran}
\bibliography{refs,more-refs,ethics-refs}

@article{nickerson1998confirmation,
 author={Nickerson, Raymond S.}, title={Confirmation Bias: A Ubiquitous Phenomenon in Many Guises},
 journal={Review of General Psychology}, year={1998}, volume={2}, number={2}, pages={175--220},
 doi={10.1037/1089-2680.2.2.175}}

@article{kunda1990case,
 author={Kunda, Ziva}, title={The Case for Motivated Reasoning}, journal={Psychological Bulletin},
 year={1990}, volume={108}, number={3}, pages={480--498}, doi={10.1037/0033-2909.108.3.480}}

@article{taber2006motivated,
 author={Taber, Charles S. and Lodge, Milton}, title={Motivated Skepticism in the Evaluation of Political Beliefs},
 journal={American Journal of Political Science}, year={2006}, volume={50}, number={3}, pages={755--769},
 doi={10.1111/j.1540-5907.2006.00214.x}}

@article{lord1979biased,
 author={Lord, Charles G. and Ross, Lee and Lepper, Mark R.},
 title={Biased Assimilation and Attitude Polarization: The Effects of Prior Theories on Subsequently Considered Evidence},
 journal={Journal of Personality and Social Psychology}, year={1979}, volume={37}, number={11}, pages={2098--2109},
 doi={10.1037/0022-3514.37.11.2098}}

@article{hart2009validated,
 author={Hart, William and Albarrac{\'i}n, Dolores and Eagly, Alice H. and Brechan, Inge and Lindberg, Matthew J. and Merrill, Lisa},
 title={Feeling Validated Versus Being Correct: A Meta-Analysis of Selective Exposure to Information},
 journal={Psychological Bulletin}, year={2009}, volume={135}, number={4}, pages={555--588}, doi={10.1037/a0015701}}

@article{stanovich2013myside,
 author={Stanovich, Keith E. and West, Richard F. and Toplak, Maggie E.},
 title={Myside Bias, Rational Thinking, and Intelligence}, journal={Current Directions in Psychological Science},
 year={2013}, volume={22}, number={4}, pages={259--264}, doi={10.1177/0963721413480174}}

@article{clifford1877ethics,
 author={Clifford, William K.}, title={The Ethics of Belief}, journal={Contemporary Review},
 year={1877}, volume={29}, pages={289--309}}

@book{medina2013epistemology,
 author={Medina, Jos{\'e}}, title={The Epistemology of Resistance: Gender and Racial Oppression, Epistemic Injustice, and the Social Imagination},
 publisher={Oxford University Press}, year={2013}, isbn={9780199929023},
 doi={10.1093/acprof:oso/9780199929023.001.0001}}

@article{feldman1988obligations,
 author={Feldman, Richard}, title={Epistemic Obligations}, journal={Philosophical Perspectives},
 year={1988}, volume={2}, pages={235--256}, doi={10.2307/2214076}}

@article{feldman2000ethics,
 author={Feldman, Richard}, title={The Ethics of Belief}, journal={Philosophy and Phenomenological Research},
 year={2000}, volume={60}, number={3}, pages={667--695}, doi={10.2307/2653823}}

@book{fricker2007epistemic,
 author={Fricker, Miranda}, title={Epistemic Injustice: Power and the Ethics of Knowing},
 publisher={Oxford University Press}, year={2007}, isbn={9780198237907},
 doi={10.1093/acprof:oso/9780198237907.001.0001}}

@techreport{solaiman2019release,
 author={Solaiman, Irene and Brundage, Miles and Clark, Jack and Askell, Amanda and Herbert-Voss, Ariel and Wu, Jeff and Radford, Alec and Krueger, Gretchen and Kim, Jong Wook and Kreps, Sarah and McCain, Miles and Newhouse, Alex and Blazakis, Jason and McGuffie, Kris and Wang, Jasmine},
 title={Release Strategies and the Social Impacts of Language Models},
 institution={OpenAI},
 year={2019},
 eprint={1908.09203}, archivePrefix={arXiv}, primaryClass={cs.CL},
 url={https://arxiv.org/abs/1908.09203}, note={arXiv:1908.09203}}

@article{risko2016offloading,
 author={Risko, Evan F. and Gilbert, Sam J.}, title={Cognitive Offloading},
 journal={Trends in Cognitive Sciences}, year={2016}, volume={20}, number={9}, pages={676--688},
 doi={10.1016/j.tics.2016.07.002}}

@article{parasuraman2010complacency,
 author={Parasuraman, Raja and Manzey, Dietrich H.},
 title={Complacency and Bias in Human Use of Automation: An Attentional Integration},
 journal={Human Factors}, year={2010}, volume={52}, number={3}, pages={381--410},
 doi={10.1177/0018720810376055}}

@inproceedings{lee2025critical,
 author={Lee, Hao-Ping (Hank) and Sarkar, Advait and Tankelevitch, Lev and Drosos, Ian and Rintel, Sean and Banks, Richard and Wilson, Nicholas},
 title={The Impact of Generative {AI} on Critical Thinking: Self-Reported Reductions in Cognitive Effort and Confidence Effects From a Survey of Knowledge Workers},
 booktitle={Proceedings of the 2025 CHI Conference on Human Factors in Computing Systems},
 publisher={Association for Computing Machinery}, year={2025}, pages={1--22}, articleno={1121},
 doi={10.1145/3706598.3713778}, note={Article 1121}}

@misc{chismon2025prompt,
 author={Chismon, Dave},
 title={Prompt Injection Is Not {SQL} Injection (It May Be Worse)},
 year={2025},
 month={December 8},
 howpublished={National Cyber Security Centre},
 url={https://www.ncsc.gov.uk/blog-post/prompt-injection-is-not-sql-injection},
 note={Accessed October 21, 2026}}

@book{wiener1950human,
 author={Wiener, Norbert},
 title={The Human Use of Human Beings: Cybernetics and Society},
 publisher={Houghton Mifflin},
 address={Boston},
 year={1950},
 note={First edition; especially Chapter XI, pp. 212--213}}

@misc{amodei2016concrete,
  author = {Amodei, Dario and Olah, Chris and Steinhardt, Jacob and Christiano, Paul and Schulman, John and Man{\'e}, Dan},
  title = {Concrete Problems in {AI} Safety},
  year = {2016},
  howpublished = {arXiv:1606.06565},
  eprint = {1606.06565},
  archivePrefix = {arXiv},
  primaryClass = {cs.AI},
  doi = {10.48550/arXiv.1606.06565},
  url = {https://arxiv.org/abs/1606.06565}
}

@misc{krakovna2020specification,
  author = {Krakovna, Victoria and Uesato, Jonathan and Mikulik, Vladimir and Rahtz, Matthew and Everitt, Tom and Kumar, Ramana and Kenton, Zac and Leike, Jan and Legg, Shane},
  title = {Specification Gaming: The Flip Side of {AI} Ingenuity},
  year = {2020},
  month = {April 21},
  howpublished = {DeepMind research blog},
  url = {https://deepmind.google/blog/specification-gaming-the-flip-side-of-ai-ingenuity/}
}

@article{rubel2019agency,
  author = {Rubel, Alan and Castro, Clinton and Pham, Adam},
  title = {Agency Laundering and Information Technologies},
  journal = {Ethical Theory and Moral Practice},
  volume = {22},
  pages = {1017--1041},
  year = {2019},
  doi = {10.1007/s10677-019-10030-w},
  url = {https://link.springer.com/article/10.1007/s10677-019-10030-w}
}

@article{bainbridge1983ironies,
  author = {Bainbridge, Lisanne},
  title = {Ironies of Automation},
  journal = {Automatica},
  volume = {19},
  number = {6},
  pages = {775--779},
  year = {1983},
  doi = {10.1016/0005-1098(83)90046-8},
  url = {https://doi.org/10.1016/0005-1098(83)90046-8}
}

@article{elish2019moral,
  author = {Elish, Madeleine Clare},
  title = {Moral Crumple Zones: Cautionary Tales in Human-Robot Interaction},
  journal = {Engaging Science, Technology, and Society},
  year = {2019},
  volume = {5},
  pages = {40--60},
  doi = {10.17351/ests2019.260},
  url = {https://estsjournal.org/index.php/ests/article/view/260}
}

@article{green2022flaws,
  author = {Green, Ben},
  title = {The Flaws of Policies Requiring Human Oversight of Government Algorithms},
  journal = {Computer Law \& Security Review},
  volume = {45},
  pages = {105681},
  year = {2022},
  doi = {10.1016/j.clsr.2022.105681},
  url = {https://doi.org/10.1016/j.clsr.2022.105681}
}

@article{santonidesio2018meaningful,
  author = {Santoni de Sio, Filippo and van den Hoven, Jeroen},
  title = {Meaningful Human Control over Autonomous Systems: A Philosophical Account},
  journal = {Frontiers in Robotics and AI},
  year = {2018},
  volume = {5},
  pages = {15},
  doi = {10.3389/frobt.2018.00015},
  url = {https://www.frontiersin.org/journals/robotics-and-ai/articles/10.3389/frobt.2018.00015/full}
}

@article{cavalcantesiebert2023meaningful,
  author = {Cavalcante Siebert, Luciano and Lupetti, Maria Luce and Aizenberg, Evgeni and Beckers, Niek and Zgonnikov, Arkady and Veluwenkamp, Herman and Abbink, David and Giaccardi, Elisa and Houben, Geert-Jan and Jonker, Catholijn M. and van den Hoven, Jeroen and Forster, Deborah and Lagendijk, Reginald L.},
  title = {Meaningful Human Control: Actionable Properties for {AI} System Development},
  journal = {AI and Ethics},
  volume = {3},
  pages = {241--255},
  year = {2023},
  doi = {10.1007/s43681-022-00167-3},
  url = {https://link.springer.com/article/10.1007/s43681-022-00167-3},
  note = {First published online in 2022; cited by the 2023 volume year}
}

@article{bucinca2021trust,
  author = {Bu{\c c}inca, Zana and Malaya, Maja Barbara and Gajos, Krzysztof Z.},
  title = {To Trust or to Think: Cognitive Forcing Functions Can Reduce Overreliance on {AI} in {AI}-Assisted Decision-Making},
  journal = {Proceedings of the ACM on Human-Computer Interaction},
  volume = {5},
  number = {CSCW1},
  eid = {188},
  note = {Article 188},
  pages = {1--21},
  year = {2021},
  doi = {10.1145/3449287},
  url = {https://doi.org/10.1145/3449287}
}

@article{leveson2004accident,
  author = {Leveson, Nancy},
  title = {A New Accident Model for Engineering Safer Systems},
  journal = {Safety Science},
  year = {2004},
  volume = {42},
  number = {4},
  pages = {237--270},
  doi = {10.1016/S0925-7535(03)00047-X},
  url = {https://doi.org/10.1016/S0925-7535(03)00047-X}
}

@article{symonshorner2020,
  author  = {Symons, John and Horner, Jack K.},
  title   = {Why There is no General Solution to the Problem of Software Verification},
  journal = {Foundations of Science},
  volume  = {25},
  number  = {3},
  pages   = {541--557},
  year    = {2020},
  doi     = {10.1007/s10699-019-09611-w},
  note    = {Published online 3 June 2019}
}

@article{symonshorner2014,
  author  = {Symons, John and Horner, Jack},
  title   = {Software Intensive Science},
  journal = {Philosophy \& Technology},
  volume  = {27},
  number  = {3},
  pages   = {461--477},
  year    = {2014},
  doi     = {10.1007/s13347-014-0163-x}
}

@article{clarke1986,
  author  = {Clarke, Edmund M. and Emerson, E. Allen and Sistla, A. Prasad},
  title   = {Automatic Verification of Finite-State Concurrent Systems Using Temporal Logic Specifications},
  journal = {ACM Transactions on Programming Languages and Systems},
  volume  = {8},
  number  = {2},
  pages   = {244--263},
  year    = {1986},
  doi     = {10.1145/5397.5399}
}

@inproceedings{klein2009,
  author    = {Klein, Gerwin and Elphinstone, Kevin and Heiser, Gernot and Andronick, June and Cock, David and Derrin, Philip and Elkaduwe, Dhammika and Engelhardt, Kai and Kolanski, Rafal and Norrish, Michael and Sewell, Thomas and Tuch, Harvey and Winwood, Simon},
  title     = {{seL4}: Formal Verification of an {OS} Kernel},
  booktitle = {Proceedings of the ACM SIGOPS 22nd Symposium on Operating Systems Principles (SOSP)},
  pages     = {207--220},
  year      = {2009},
  doi       = {10.1145/1629575.1629596}
}

@inproceedings{cousot1977,
  author    = {Cousot, Patrick and Cousot, Radhia},
  title     = {Abstract Interpretation: A Unified Lattice Model for Static Analysis of Programs by Construction or Approximation of Fixpoints},
  booktitle = {Conference Record of the 4th ACM SIGACT-SIGPLAN Symposium on Principles of Programming Languages (POPL)},
  pages     = {238--252},
  year      = {1977},
  doi       = {10.1145/512950.512973}
}

@article{merrillsabharwal2023,
  author  = {Merrill, William and Sabharwal, Ashish},
  title   = {The Parallelism Tradeoff: Limitations of Log-Precision Transformers},
  journal = {Transactions of the Association for Computational Linguistics},
  volume  = {11},
  pages   = {531--545},
  year    = {2023},
  doi     = {10.1162/tacl_a_00562}
}

@article{merrill2022,
  author  = {Merrill, William and Sabharwal, Ashish and Smith, Noah A.},
  title   = {Saturated Transformers are Constant-Depth Threshold Circuits},
  journal = {Transactions of the Association for Computational Linguistics},
  volume  = {10},
  pages   = {843--856},
  year    = {2022},
  doi     = {10.1162/tacl_a_00493}
}

@misc{schuurmans2023,
  author       = {Schuurmans, Dale},
  title        = {Memory Augmented Large Language Models are Computationally Universal},
  howpublished = {arXiv:2301.04589},
  year         = {2023},
  doi          = {10.48550/arXiv.2301.04589}
}

@article{perezbarcelo2021,
  author  = {P{\'e}rez, Jorge and Barcel{\'o}, Pablo and Marinkovic, Javier},
  title   = {Attention is {T}uring Complete},
  journal = {Journal of Machine Learning Research},
  volume  = {22},
  number  = {75},
  pages   = {1--35},
  year    = {2021}
}

@inproceedings{shacham2007,
  author    = {Shacham, Hovav},
  title     = {The Geometry of Innocent Flesh on the Bone: Return-into-libc without Function Calls (on the x86)},
  booktitle = {Proceedings of the 14th ACM Conference on Computer and Communications Security (CCS)},
  pages     = {552--561},
  year      = {2007},
  doi       = {10.1145/1315245.1315313}
}

@article{dullien2020,
  author  = {Dullien, Thomas},
  title   = {Weird Machines, Exploitability, and Provable Unexploitability},
  journal = {IEEE Transactions on Emerging Topics in Computing},
  volume  = {8},
  number  = {2},
  pages   = {391--403},
  year    = {2020},
  doi     = {10.1109/TETC.2017.2785299}
}

@article{lampson1973,
  author  = {Lampson, Butler W.},
  title   = {A Note on the Confinement Problem},
  journal = {Communications of the ACM},
  volume  = {16},
  number  = {10},
  pages   = {613--615},
  year    = {1973},
  doi     = {10.1145/362375.362389}
}

@inproceedings{kocher2019,
  author    = {Kocher, Paul and Horn, Jann and Fogh, Anders and Genkin, Daniel and Gruss, Daniel and Haas, Werner and Hamburg, Mike and Lipp, Moritz and Mangard, Stefan and Prescher, Thomas and Schwarz, Michael and Yarom, Yuval},
  title     = {Spectre Attacks: Exploiting Speculative Execution},
  booktitle = {2019 IEEE Symposium on Security and Privacy (SP)},
  pages     = {1--19},
  year      = {2019},
  doi       = {10.1109/SP.2019.00002}
}

@inproceedings{sharma2024,
  author    = {Sharma, Mrinank and Tong, Meg and Korbak, Tomasz and Duvenaud, David and Askell, Amanda and Bowman, Samuel R. and Cheng, Newton and Durmus, Esin and Hatfield-Dodds, Zac and Johnston, Scott R. and Kravec, Shauna and Maxwell, Timothy and McCandlish, Sam and Ndousse, Kamal and Rausch, Oliver and Schiefer, Nicholas and Yan, Da and Zhang, Miranda and Perez, Ethan},
  title     = {Towards Understanding Sycophancy in Language Models},
  booktitle = {The Twelfth International Conference on Learning Representations (ICLR)},
  year      = {2024},
  note      = {Preprint: arXiv:2310.13548, 2023}
}

@misc{weidinger2021,
  author       = {Weidinger, Laura and Mellor, John and Rauh, Maribeth and Griffin, Conor and Uesato, Jonathan and Huang, Po-Sen and Cheng, Myra and Glaese, Mia and Balle, Borja and Kasirzadeh, Atoosa and Kenton, Zac and Brown, Sasha and Hawkins, Will and Stepleton, Tom and Biles, Courtney and Birhane, Abeba and Haas, Julia and Rimell, Laura and Hendricks, Lisa Anne and Isaac, William and Legassick, Sean and Irving, Geoffrey and Gabriel, Iason},
  title        = {Ethical and Social Risks of Harm from Language Models},
  howpublished = {arXiv:2112.04359},
  year         = {2021},
  doi          = {10.48550/arXiv.2112.04359}
}

@article{salvi2025,
  author  = {Salvi, Francesco and Horta Ribeiro, Manoel and Gallotti, Riccardo and West, Robert},
  title   = {On the Conversational Persuasiveness of {GPT-4}},
  journal = {Nature Human Behaviour},
  volume  = {9},
  number  = {8},
  pages   = {1645--1653},
  year    = {2025},
  doi     = {10.1038/s41562-025-02194-6},
  note    = {Preprint: ``On the Conversational Persuasiveness of Large Language Models: A Randomized Controlled Trial,'' arXiv:2403.14380, 2024}
}

@inproceedings{markstegmark2023,
  author       = {Marks, Samuel and Tegmark, Max},
  title        = {The Geometry of Truth: Emergent Linear Structure in {LLM} Representations of True/False Datasets},
  booktitle    = {First Conference on Language Modeling (COLM)},
  year         = {2024},
  url          = {https://arxiv.org/abs/2310.06824},
  note         = {Preprint: arXiv:2310.06824, 2023}
}

@misc{zou2023,
  author       = {Zou, Andy and Phan, Long and Chen, Sarah and Campbell, James and Guo, Phillip and Ren, Richard and others},
  title        = {Representation Engineering: A Top-Down Approach to {AI} Transparency},
  howpublished = {arXiv:2310.01405},
  year         = {2023},
  doi          = {10.48550/arXiv.2310.01405}
}

@article{levinsteinherrmann2024,
  author  = {Levinstein, Benjamin A. and Herrmann, Daniel A.},
  title   = {Still No Lie Detector for Language Models: Probing Empirical and Conceptual Roadblocks},
  journal = {Philosophical Studies},
  volume  = {182},
  pages   = {1539--1565},
  year    = {2025},
  doi     = {10.1007/s11098-023-02094-3},
  note    = {Published online 17 February 2024; preprint: arXiv:2307.00175, 2023}
}

@inproceedings{adleman1990,
  author    = {Adleman, Leonard M.},
  title     = {An Abstract Theory of Computer Viruses},
  booktitle = {Advances in Cryptology --- {CRYPTO} '88},
  series    = {Lecture Notes in Computer Science},
  volume    = {403},
  pages     = {354--374},
  publisher = {Springer},
  year      = {1990},
  doi       = {10.1007/0-387-34799-2_28}
}

@article{alfonseca2021,
  author  = {Alfonseca, Manuel and Cebrian, Manuel and Fern{\'a}ndez Anta, Antonio and Coviello, Lorenzo and Abeliuk, Andr{\'e}s and Rahwan, Iyad},
  title   = {Superintelligence Cannot be Contained: Lessons from Computability Theory},
  journal = {Journal of Artificial Intelligence Research},
  volume  = {70},
  pages   = {65--76},
  year    = {2021},
  doi     = {10.1613/jair.1.12202},
  note    = {Preprint: arXiv:1607.00913, 2016}
}

@inproceedings{ball2026,
  author    = {Ball, Sarah and Gluch, Greg and Goldwasser, Shafi and Kreuter, Frauke and Reingold, Omer and Rothblum, Guy N.},
  title     = {On the Impossibility of Separating Intelligence from Judgment: The Computational Intractability of Filtering for {AI} Alignment},
  booktitle = {The Fourteenth International Conference on Learning Representations ({ICLR})},
  year      = {2026},
  note      = {Earlier preprint: ``The Computational Intractability of Filtering for {AI} Alignment,'' arXiv:2507.07341, 2025}
}

@incollection{deleeuw1956,
  author    = {de Leeuw, K. and Moore, E. F. and Shannon, C. E. and Shapiro, N.},
  title     = {Computability by Probabilistic Machines},
  booktitle = {Automata Studies},
  editor    = {Shannon, C. E. and McCarthy, J.},
  series    = {Annals of Mathematics Studies},
  volume    = {34},
  pages     = {183--212},
  publisher = {Princeton University Press},
  year      = {1956}
}

@inproceedings{kalai2024,
  author    = {Kalai, Adam Tauman and Vempala, Santosh S.},
  title     = {Calibrated Language Models Must Hallucinate},
  booktitle = {Proceedings of the 56th Annual {ACM} Symposium on Theory of Computing ({STOC})},
  pages     = {160--171},
  year      = {2024},
  doi       = {10.1145/3618260.3649777},
  note      = {Preprint: arXiv:2311.14648, 2023}
}

@article{ligatti2005,
  author  = {Ligatti, Jay and Bauer, Lujo and Walker, David},
  title   = {Edit Automata: Enforcement Mechanisms for Run-time Security Policies},
  journal = {International Journal of Information Security},
  volume  = {4},
  number  = {1--2},
  pages   = {2--16},
  year    = {2005}
}

@phdthesis{viswanathan2000,
  author = {Viswanathan, Mahesh},
  title  = {Foundations for the Run-Time Analysis of Software Systems},
  school = {University of Pennsylvania},
  year   = {2000}
}

@misc{long2026guardrail,
  author       = {Long, Darrell},
  title        = {{No Perfect Guardrail}},
  howpublished = {Blog post},
  year         = {2026},
  month        = sep # " 29",
  note         = {Accessed November 19, 2026},
  url          = {https://darrelllong.github.io/blog/2026-09-29-no-perfect-guardrail/}
}

@article{turing1936,
  author  = {Turing, Alan M.},
  title   = {On Computable Numbers, with an Application to the {E}ntscheidungsproblem},
  journal = {Proceedings of the London Mathematical Society},
  volume  = {s2-42},
  number  = {1},
  pages   = {230--265},
  year    = {1937},
  note    = {Received 28 May 1936},
  doi = {10.1112/plms/s2-42.1.230}
}

@article{rice1953,
  author  = {Rice, H. Gordon},
  title   = {Classes of Recursively Enumerable Sets and Their Decision Problems},
  journal = {Transactions of the American Mathematical Society},
  volume  = {74},
  number  = {2},
  pages   = {358--366},
  year    = {1953},
  doi = {10.1090/S0002-9947-1953-0053041-6}
}

@article{chomsky1959,
  author  = {Chomsky, Noam},
  title   = {On Certain Formal Properties of Grammars},
  journal = {Information and Control},
  volume  = {2},
  number  = {2},
  pages   = {137--167},
  year    = {1959},
  doi = {10.1016/S0019-9958(59)90362-6}
}

@book{rogers1967,
  author    = {Rogers, Jr., Hartley},
  title     = {Theory of Recursive Functions and Effective Computability},
  publisher = {McGraw-Hill},
  address   = {New York},
  year      = {1967}
}

@book{hopcroftullman1979,
  author    = {Hopcroft, John E. and Ullman, Jeffrey D.},
  title     = {Introduction to Automata Theory, Languages, and Computation},
  publisher = {Addison-Wesley},
  address   = {Reading, MA},
  year      = {1979}
}

@article{cohen1987,
  author  = {Cohen, Fred},
  title   = {Computer Viruses: Theory and Experiments},
  journal = {Computers \& Security},
  volume  = {6},
  number  = {1},
  pages   = {22--35},
  year    = {1987}
}

@article{schneider2000,
  author  = {Schneider, Fred B.},
  title   = {Enforceable Security Policies},
  journal = {ACM Transactions on Information and System Security},
  volume  = {3},
  number  = {1},
  pages   = {30--50},
  year    = {2000},
  doi = {10.1145/353323.353382}
}

@inproceedings{glukhov2023,
  author = {Glukhov, David and Shumailov, Ilia and Gal, Yarin and Papernot, Nicolas and Papyan, Vardan},
  title = {Position: Fundamental Limitations of {LLM} Censorship Necessitate New Approaches},
  booktitle = {Proceedings of the 41st International Conference on Machine Learning},
  series = {Proceedings of Machine Learning Research},
  volume = {235},
  pages = {15767--15787},
  publisher = {PMLR},
  year = {2024},
  url = {https://proceedings.mlr.press/v235/glukhov24a.html},
  note = {Earlier preprint: ``LLM Censorship: A Machine Learning Challenge or a Computer Security Problem?,'' arXiv:2307.10719, 2023}
}

@misc{xu2024,
  author       = {Xu, Ziwei and Jain, Sanjay and Kankanhalli, Mohan},
  title        = {Hallucination is Inevitable: An Innate Limitation of Large Language Models},
  howpublished = {arXiv:2401.11817},
  year         = {2024},
  url = {https://arxiv.org/abs/2401.11817},
  note = {Revised February 13, 2025 (arXiv:2401.11817v2)}
}

@article{melo2024,
  author  = {Melo, Gabriel A. and M{\'a}ximo, Marcos R. O. A. and Soma, Nei Y. and Castro, Paulo A. L.},
  title   = {Machines that Halt Resolve the Undecidability of Artificial Intelligence Alignment},
  journal = {Scientific Reports},
  volume  = {15},
  pages   = {15591},
  year    = {2025},
  note    = {Preprint: arXiv:2408.08995, 2024},
  doi = {10.1038/s41598-025-99060-2}
}

@incollection{presburger1929,
  author    = {Presburger, Moj{\.z}esz},
  title     = {{\"U}ber die {V}ollst{\"a}ndigkeit eines gewissen {S}ystems der {A}rithmetik ganzer {Z}ahlen, in welchem die {A}ddition als einzige {O}peration hervortritt},
  booktitle = {Comptes Rendus du I Congr{\`e}s des Math{\'e}maticiens des Pays Slaves, Warszawa 1929},
  editor    = {Leja, F.},
  pages     = {92--101, 395},
  publisher = {Ksi{\k{a}}{\.z}nica Atlas},
  address   = {Warsaw},
  year      = {1930}
}

@article{hamlen2006,
  author  = {Hamlen, Kevin W. and Morrisett, Greg and Schneider, Fred B.},
  title   = {Computability Classes for Enforcement Mechanisms},
  journal = {ACM Transactions on Programming Languages and Systems},
  volume  = {28},
  number  = {1},
  pages   = {175--205},
  year    = {2006},
  doi = {10.1145/1111596.1111601}
}

@article{godel1931,
  author  = {G{\"o}del, Kurt},
  title   = {{\"U}ber formal unentscheidbare {S}{\"a}tze der {P}rincipia {M}athematica und verwandter {S}ysteme {I}},
  journal = {Monatshefte f{\"u}r Mathematik und Physik},
  volume  = {38},
  pages   = {173--198},
  year    = {1931},
  doi = {10.1007/BF01700692}
}

\newpage
\appendix
\renewcommand{\thesection}{Appendix}
\renewcommand{\thesubsection}{\arabic{subsection}}
\section{Fuller Proof and Further Results}\label{app:proofs}

In this appendix, we present the complete formal development in one place. We
repeat the notation, definitions, assumptions, results, and proofs needed to
read the argument independently of the main text, including the qualifications
on bounded outputs, average error, adversaries, and restricted programs.
The central guard theorem and the further guard results are stated separately,
as in the main text, but all their proofs are included here.

For comparison with the body, Theorem~\ref{app:full:thm:general} restates
Theorem~\ref{thm:general}; Theorems~\ref{app:full:thm:truth},
\ref{app:full:thm:rice-traces}, and~\ref{app:full:thm:monitor} restate
Theorems~\ref{thm:truth}, \ref{thm:rice-traces}, and~\ref{thm:monitor},
respectively. The repeated definitions, corollaries, and delivery proposition
also retain their substance. Appendix results use an A prefix to distinguish
them from the results in the body.

\setcounter{theorem}{0}
\setcounter{proposition}{0}
\setcounter{corollary}{0}
\setcounter{definition}{0}
\renewcommand{\thetheorem}{A\arabic{theorem}}
\renewcommand{\theproposition}{A\arabic{proposition}}
\renewcommand{\thecorollary}{A\arabic{corollary}}
\renewcommand{\thedefinition}{A\arabic{definition}}
\renewcommand{\theHtheorem}{appendix.\arabic{theorem}}
\renewcommand{\theHproposition}{appendix.\arabic{proposition}}
\renewcommand{\theHcorollary}{appendix.\arabic{corollary}}
\renewcommand{\theHdefinition}{appendix.\arabic{definition}}

\subsection{Notation and standard facts}\label{app:facts}

\paragraph{Notation.}
$\Sigma$ is a finite alphabet with at least two symbols, $\Str$ is the set of
finite strings over $\Sigma$, $\emptystr$ is the empty string, $|w|$ is the
length of the string $w$, and $\Sigma^{\le n}$ is the set of strings of length
at most $n$.  A \emph{language} is a subset of $\Str$.  The \emph{complement}
of a language $B$ is $\co B=\Str\setminus B$, and the \emph{symmetric
difference} of languages $B$ and $C$ is
$B\symdiff C=(B\setminus C)\cup(C\setminus B)$.  We write
$\N=\{0,1,2,\dots\}$.

We assume the definition of a Turing
machine~\cite{hopcroftullman1979}, and fix a standard coding under which every
string $e$ is the code of a machine $M_e$~\cite{rogers1967}.  We write
$M_e(x)\halts$ if $M_e$ halts on input $x$, and $M_e(x)\diverges$ if it does
not.  A function is \emph{computable} if some Turing machine halts on every
input $x$ with output $f(x)$; arguments and values that are not strings are
coded as strings.  A language $B$ is \emph{decidable} if the function that is
$1$ on $B$ and $0$ on $\co B$ is computable, and \emph{undecidable} otherwise.
It is \emph{recursively enumerable} if it is the set of inputs on which some
Turing machine halts.  A language $B$ is \emph{many-one reducible} to a
language $C$, written $B\mred C$, if there is a computable function $f$ such
that $x\in B$ if and only if $f(x)\in C$, for every $x$.  The \emph{halting
set} is $K=\{e:M_e(e)\halts\}$.  We use three standard
facts~\cite{turing1936,rogers1967}.
\begin{enumerate}
\item[(F1)] $K$ is recursively enumerable and undecidable.
\item[(F2)] A language is decidable if and only if it and its complement are
  both recursively enumerable.
\item[(F3)] If $K\mred C$, then $C$ is undecidable and $\co C$ is not
  recursively enumerable.
\end{enumerate}

\subsection{Generators, policies, and guards}\label{app:full:model}

\begin{definition}[Generator, policy, guard]\label{app:full:def:guard}
A \emph{generator} is a computable function $G:\Str\times\Str\to\Str$.  The
string $G(u,r)$ is the output produced in response to the \emph{prompt} $u$
when the random choices of the generator are determined by the \emph{seed}
$r$.  The set of possible outputs on prompt $u$ is
$\Out_G(u)=\{G(u,r):r\in\Str\}$, and $\Out_G$ is the union of these sets over
all prompts.  The generator is \emph{full} if $\Out_G=\Str$.

A \emph{policy} is a language $B$, the set of \emph{forbidden} outputs.
Outputs in $\co B$ are \emph{permitted}.  A \emph{guard} is a computable
function $F:\Str\times\Str\to\{0,1\}$.  It \emph{blocks} the output $w$ on
prompt $u$ if $F(u,w)=1$ and \emph{passes} it if $F(u,w)=0$.  With respect to
$G$ and $B$, a guard is \emph{sound} if it blocks every forbidden output of
$G$ on every prompt, \emph{permissive} if it passes every permitted output of
$G$ on every prompt, and \emph{exact} if it is both.

A \emph{randomised guard} is a Turing machine that may toss a fair coin at any
step and that, on every input $(u,w)$ and every sequence of outcomes of its
tosses, halts with output $0$ or $1$.
\end{definition}

A guard is the formal counterpart of a guardrail, and an exact guard is a
perfect one.  The guard that blocks every output is sound, and the guard that
passes every output is permissive.  The question is whether one guard can be
both.  We take each policy as a given set; how a policy is specified is a
separate question.  A generator models a generation procedure that terminates
on every prompt and seed.  A guard may itself be a language model.

\subsection{The central guard theorem}\label{app:full:guards}

\begin{theorem}\label{app:full:thm:general}
Let $G$ be a full generator and $B$ a policy.
\begin{enumerate}
\item[\textup{(a)}] An exact guard exists if and only if $B$ is decidable.
\item[\textup{(b)}] If $B$ is undecidable, every guard gives the wrong verdict
  on infinitely many outputs; every sound guard blocks infinitely many
  permitted outputs; and every permissive guard passes infinitely many
  forbidden outputs.
\end{enumerate}
\end{theorem}

\begin{proof}
Since $\Out_G=\Str$, the following search halts for every string $w$: list the
pairs $(u,r)$ in a fixed order and stop at the first with $G(u,r)=w$.  Let
$\pi(w)$ be the prompt of that pair.  Then $\pi$ is computable and
$w\in\Out_G(\pi(w))$.  For a guard $F$ let $D_F=\{w:F(\pi(w),w)=0\}$, which is
decidable.  If $F$ is sound then $D_F\subseteq\co B$, and if $F$ is permissive
then $\co B\subseteq D_F$.

(a)~If $B$ is decidable, the guard that blocks $w$ if and only if $w\in B$ is
exact.  If $F$ is exact then $D_F=\co B$, so $B$ is decidable.

(b)~Let $E=D_F\symdiff\co B$, so that $\co B=D_F\symdiff E$.  If $E$ were
finite it would be decidable, and then $\co B$ would be decidable.  So $E$ is
infinite.  On the prompt $\pi(w)$, each $w\in E$ is either forbidden and
passed, or permitted and blocked.  If $F$ is sound then $E=\co B\setminus D_F$,
and if $F$ is permissive then $E=D_F\cap B$.

\end{proof}

The theorem has the following limits.  To conclude that no exact guard exists,
the generator need not be full: it suffices that $K$ is many-one reducible to
$B$ or to $\co B$ by a function whose every value the generator can emit.
Part~(b) counts strings and does not bound an average error
(Section~\ref{app:full:sec:limits}).  Theorem~\ref{thm:further}(a) does not say that no guard can be proved
sound; the guard that blocks every string is sound.  Theorem~\ref{thm:further}(b) requires that
the randomised guard halt on every sequence of tosses.

\subsection{Further results on guards}\label{app:more}

\paragraph{Bounded outputs.}\label{app:full:sec:finite}
A language model that stops at a limit of length is a generator, and it is not
full.  If the outputs of a generator are in $\Sigma^{\le n}$, the set of its
forbidden outputs is finite, so it is decidable, and an exact guard exists.
Theorem~\ref{app:full:thm:general} does not apply.

\begin{proposition}\label{prop:finite}
Let $B$ be an undecidable policy.  For every $n$ the language
$B\cap\Sigma^{\le n}$ is decidable.  No algorithm produces, from $n$, the code
of a machine that decides $B\cap\Sigma^{\le n}$.
\end{proposition}

\begin{proof}
The language is finite.  Given such an algorithm, one would decide whether
$w\in B$ by producing the machine for $n=|w|$ and running it on $w$.
\end{proof}

The proposition does not say that a particular bounded system lacks an exact
guard, or that its exact guard is costly.  Fullness is an additional
assumption about which outputs the generator can produce; allowing unlimited
memory does not establish it.

\begin{theorem}\label{thm:further}
Let $G$ be a full generator and $B$ a policy.
\begin{enumerate}
\item[\textup{(a)}] If $B$ is infinite, no algorithm decides, given the code
  of a machine that computes a guard, whether the guard is sound.
\item[\textup{(b)}] If $B$ is undecidable, for every randomised guard there is
  an output on which it gives the wrong verdict with probability at least
  $\tfrac12$.
\item[\textup{(c)}] If $\co B$ is not recursively enumerable, every generator
  all of whose outputs are permitted, whether it is full or not, fails to emit
  infinitely many permitted strings.
\end{enumerate}
\end{theorem}

\begin{proof}
(a)~For each $e$ let $F_e$ be the guard that, on every prompt, passes $w$ if
and only if $M_e$ halts on input $e$ within $|w|$ steps.  A code of a machine
that computes $F_e$ is computable from $e$.  If $e\notin K$, then $F_e$ blocks
every string and is sound.  If $e\in K$, then $M_e$ halts on $e$ within some
number $t$ of steps, and $F_e$ passes every string of length at least $t$.
Since $B$ is infinite, one of these strings is forbidden, and it is an output
of $G$; so $F_e$ is not sound.  An algorithm that decided soundness would
decide $K$.

(b)~Let $R$ be a randomised guard and fix $(u,w)$.  The computations of $R$ on
$(u,w)$ form a tree that branches in two at each toss and has no infinite
branch.  By K\"onig's lemma, which states that a tree with finitely many
successors at every node and no infinite branch is finite, the tree is finite.
So the probability $b(u,w)$ that $R$ outputs $1$ on $(u,w)$ is a rational
number computable from $(u,w)$.  Suppose that $R$ gave the wrong verdict with
probability less than $\tfrac12$ on every output.  Then $b(u,w)>\tfrac12$ for
every forbidden output $w$ on prompt $u$, and $b(u,w)<\tfrac12$ for every
permitted one.  The guard that blocks $w$ on prompt $u$ if and only if
$b(u,w)>\tfrac12$ would then be exact, contrary to Theorem~\ref{app:full:thm:general}(a).

(c)~Let $G'$ be a generator with $\Out_{G'}\subseteq\co B$.  The set
$\Out_{G'}$ is recursively enumerable: search for a pair $(u,r)$ with
$G'(u,r)=w$.  If $\co B\setminus\Out_{G'}$ were finite, $\co B$ would be the
union of a recursively enumerable language and a finite one, and so would be
recursively enumerable.
\end{proof}

Part~(a) does not say that no guard can be proved sound; the guard that blocks
every string is sound.  Part~(b) requires that the randomised guard halt on
every sequence of tosses.

\subsection{Policies}\label{app:full:sec:policies}

\subsubsection{Banned phrasings}\label{app:full:sec:lexical}

Let $L$ be a regular language of banned phrasings, and let $B_L$ be the set of
strings in which some member of $L$ occurs as a contiguous part.  Then $B_L$
is regular, and a finite automaton decides whether $w\in B_L$ in $|w|$
steps~\cite{hopcroftullman1979}.  By Theorem~\ref{app:full:thm:general}(a) an exact
guard exists.  A finite list of banned phrases is a special case.

Regular languages are Type~3 of the hierarchy of
Chomsky~\cite{chomsky1959}.  A list of banned words takes no account of
context.  A regular language can take some: the set of strings in which a
given word occurs without a second given word immediately after it is regular.
Other dependence on context is not regular.  The set of strings over an
opening and a closing bracket in which the brackets are correctly nested is
context-free (Type~2) and not regular.  Every context-free language is
decidable~\cite{hopcroftullman1979}.

\subsubsection{False sentences of arithmetic}\label{app:full:sec:truth}

A \emph{sentence of arithmetic} is a sentence of the first-order language with
the symbols $0$, $S$ (successor), $+$, $\times$, and $<$.  It is \emph{true}
if it holds in $\N$.  Fix a standard effective G\"odel coding of formulas by
strings~\cite{godel1931}.  Let $\Sent$ be the set of strings that code
sentences, and $\Th\subseteq\Sent$ the set of strings that code true
sentences.  $\Sent$ is decidable, and the code of the negation of a sentence
is computable from the code of the sentence.  A policy $B$ is a
\emph{falsehood policy} if $B\cap\Sent=\Sent\setminus\Th$, that is, if among
the sentences of arithmetic it forbids every false one and no true one.

\begin{theorem}\label{app:full:thm:truth}
No falsehood policy is decidable, and none is recursively enumerable or has a
recursively enumerable complement.
\end{theorem}

\begin{proof}
By G\"odel's first incompleteness theorem~\cite{godel1931}, $\Th$ is not
recursively enumerable: if it were, the true sentences would be a set of
axioms that can be listed by a machine, that is consistent, that includes the
axioms of arithmetic, and that settles every sentence, and the theorem states
that there is no such set.  Let $B$ be a falsehood policy.  If $\co B$ were
recursively enumerable, so would be $\Th=\Sent\cap\co B$.  If $B$ were
recursively enumerable, so would be the set $\Sent\cap B$ of false sentences;
then $\Th$ would be recursively enumerable, because a sentence is true if and
only if its negation is false.  By (F2) a decidable language is recursively
enumerable.
\end{proof}

The theorem concerns a policy that classifies every sentence of arithmetic
correctly.  It does not say that every policy about false statements is
undecidable: truth in the first-order arithmetic of addition alone is
decidable~\cite{presburger1929}.  For Theorem~\ref{app:full:thm:general} to apply it
suffices that the generator can emit the code of every sentence.  By
Theorem~\ref{thm:further}(c), a generator that emits only true sentences of
arithmetic fails to emit infinitely many true sentences.

\subsubsection{Programs}\label{app:full:sec:programs}

An output that is run as a program has effects while it runs, and not only a
final result.  Fix a finite non-empty alphabet $A$ of \emph{actions}, such as
commands to a device, calls to an operating system, and symbols written to a
file.  Let $A^{*}$ be the set of finite sequences of actions, $A^{\omega}$ the
set of infinite sequences, and $\Traces=A^{*}\cup A^{\omega}$.

\begin{definition}[Action machine, trace, behavior]\label{app:full:def:action}
An \emph{action machine} is a deterministic Turing machine with an input tape,
work tapes, and a write-only action tape over $A$ on which the head moves only
to the right.  To write a symbol on the action tape is to perform that action.
Fix a standard coding under which every string $p$ is the code of an action
machine $N_p$.  We call $p$ a \emph{program}.  The \emph{trace} $\tr_p(x)$ is
the finite or infinite sequence of actions that $N_p$ performs on input $x$,
in order.  It is defined whether or not $N_p$ halts.  The \emph{behavior} of
$p$ is the function $\beta_p:\Str\to\Traces$ that maps $x$ to $\tr_p(x)$.  The
\emph{silent behavior} is the behavior that performs no action on any input.
A policy $B$ is \emph{behavioral} if $\beta_p=\beta_{p'}$ implies that $p$
and $p'$ are both in $B$ or both outside it.
\end{definition}

This is a simplified trace-based model of programs, following Hamlen et
al.~\cite{hamlen2006}.  A behavioral policy judges a program by the actions
it performs and by nothing else.  A policy that forbids programs that perform
a given action within a given number of steps is not behavioral, and it is
decidable by running the program for that number of steps.

\begin{theorem}[Rice's theorem for traces]\label{app:full:thm:rice-traces}
Let $B$ be a behavioral policy that is neither empty nor all of $\Str$.  If
no program with the silent behavior is in $B$, then $K\mred B$.  Otherwise
$K\mred\co B$.  In both cases $B$ is undecidable.
\end{theorem}

\begin{proof}
Suppose that no program with the silent behavior is in $B$, and let $q\in B$.
For each $e$ let $f(e)$ be a code of the action machine that, on input $x$,
first simulates $M_e$ on input $e$, performing no action, and then, if that
simulation halts, performs the actions that $N_q$ performs on $x$.  The
function $f$ is computable.  If $e\in K$ then $\beta_{f(e)}=\beta_q$, so
$f(e)\in B$.  If $e\notin K$ then $f(e)$ has the silent behavior, so
$f(e)\notin B$.  Hence $K\mred B$.  Otherwise every program with the silent
behavior is in $B$; apply the argument to $\co B$, which is behavioral.
Undecidability follows by (F3).
\end{proof}

Rice's theorem~\cite{rice1953} is the same statement for Turing machines that
are judged by the result they return on each input on which they halt.  As
stated it does not mention actions, and a program can perform a harmful action
and never return a result.  We therefore state the theorem for traces.  The
proof is that of Rice's theorem.  It applies to traces because the machine
with code $f(e)$ performs no action while it waits.

\begin{corollary}\label{app:full:cor:harm}
Let $H\subseteq A$ be a non-empty set of harmful actions, let $x_0\in\Str$,
and let $\Harm_H$ be the set of programs $p$ such that an action in $H$ occurs
in $\tr_p(x_0)$.  Then $\Harm_H$ is recursively enumerable, $K\mred\Harm_H$,
$\Harm_H$ is undecidable, and its complement, the set of safe programs, is not
recursively enumerable.  The same holds for the set of programs that perform
an action in $H$ on some input.
\end{corollary}

\begin{proof}
$\Harm_H$ is behavioral.  It contains no program with the silent behavior,
and it contains a program that performs an action in $H$ on every input.
Apply Theorem~\ref{app:full:thm:rice-traces} and (F3).  A machine that runs $N_p$ on
$x_0$ and halts when an action in $H$ is performed halts exactly on the
members of $\Harm_H$.  For the last statement, the machine runs $N_p$ on all
inputs in turn, for increasing numbers of steps.
\end{proof}

\subsubsection{An idealized recipient}\label{app:full:sec:persons}

An output may do harm through whoever receives it.  A \emph{recipient} is a
function $\rho:\Str\to\Traces$, where $\rho(w)$ is the sequence of actions
that the recipient performs upon receiving the message $w$.  The recipient is
not assumed to be the behavior of any program.  For a non-empty set
$H\subseteq A$ of harmful actions, the policy \emph{induced} by $\rho$ is the
set $B_H(\rho)$ of messages $w$ such that an action in $H$ occurs in
$\rho(w)$.  A recipient is \emph{instructable} if there is a computable
function $\iota$ such that $\rho(\iota(p))=\tr_p(\emptystr)$ for every program
$p$.  The message $\iota(p)$ instructs the recipient to carry out the program
$p$, and the recipient does so exactly.

\begin{corollary}\label{app:full:cor:recipient}
If $\rho$ is instructable, then $K\mred B_H(\rho)$.  Hence $B_H(\rho)$ is
undecidable and the set of harmless messages is not recursively enumerable.
If in addition $\rho$ is the behavior $\beta_c$ of some program $c$, then
$B_H(\rho)$ is recursively enumerable.
\end{corollary}

\begin{proof}
Let $f$ be the reduction of $K$ to $\Harm_H$ with $x_0=\emptystr$ given by
Theorem~\ref{app:full:thm:rice-traces}.  Then $e\in K$ if and only if
$\iota(f(e))\in B_H(\rho)$.  Apply (F3).  If $\rho=\beta_c$, a machine that
runs $N_c$ on $w$ and halts when an action in $H$ is performed halts exactly
on the members of $B_H(\rho)$.
\end{proof}

An instructable recipient is an interpreter of programs that is exact and
unlimited in time and memory.  The corollary applies to a person only if there
is an encoding $\iota$ of instructions that the person carries out in this
way.  It is not a theorem about persuasion.


\subsection{Monitors}\label{app:full:sec:monitors}

A guard examines a program before it is run.  A monitor examines the actions
of a program while it runs, and can stop the program.

\begin{definition}[Trace policy, monitor]\label{app:full:def:monitor}
A \emph{trace policy} is a set $\Bad\subseteq A^{*}$ that does not contain the
empty sequence and that contains $sz$ whenever it contains $s$.  A finite
trace is \emph{forbidden} if it is in $\Bad$ and \emph{permitted} otherwise.
A \emph{monitor} is a computable function $\mu:A^{*}\to\{0,1\}$.  When a
program run under the monitor has performed the sequence $s$ and is about to
perform the action $a$, the program is stopped if $\mu(sa)=1$, and $a$ is
performed if $\mu(sa)=0$.  A monitor is \emph{sound} if $\mu(sa)=1$ whenever
$s$ is permitted and $sa$ is forbidden, \emph{permissive} if $\mu(s)=0$
whenever $s$ is permitted, and \emph{exact} if it is both.
\end{definition}

\begin{theorem}\label{app:full:thm:monitor}
An exact monitor exists if and only if $\Bad$ is decidable.  If $\Bad$ is
undecidable, then under every sound monitor there are infinitely many
permitted finite traces that cannot be performed in full, and under every
permissive monitor there are infinitely many forbidden finite traces that can.
\end{theorem}

\begin{proof}
Let $D_\mu$ be the set of finite traces $s$ such that $\mu$ is $0$ on every
non-empty initial segment of $s$.  These are the traces that can be performed
in full.  $D_\mu$ is decidable.  If $\mu$ is sound, no forbidden trace is in
$D_\mu$, because $\mu$ is $1$ on its shortest forbidden initial segment.  If
$\mu$ is permissive, every permitted trace is in $D_\mu$, because all its
initial segments are permitted.  So if $\mu$ is exact, $D_\mu$ is the
complement of $\Bad$, and $\Bad$ is decidable.  Conversely, if $\Bad$ is
decidable, the function that is $1$ on $\Bad$ and $0$ elsewhere is an exact
monitor.  The rest follows as in Theorem~\ref{app:full:thm:general}(b).
\end{proof}

A trace policy corresponds to a safety property in the sense of
Schneider~\cite{schneider2000}, who showed that a monitor that can only stop a
program enforces no other kind of policy.  The condition that $\Bad$ be
decidable corresponds to the condition that Hamlen et
al.~\cite{hamlen2006} place on the test that a monitor applies to finite
sequences of events.  Whether it holds depends on what the monitor observes
and on what counts as the violation.  There are two cases.

\paragraph{The violation is an action.}
Let $H\subseteq A$ be a non-empty set of harmful actions, and let $\Bad$ be
the set of finite traces in which an action in $H$ occurs.  This set is
decidable.  The monitor that stops a program exactly when its next action is
in $H$ is exact.  By Corollary~\ref{app:full:cor:harm} no guard decides from the text
of a program whether the program will perform an action in $H$.  The monitor
prevents every such action and stops no program whose trace is permitted.
This requires that the harmful actions be specified as a set of symbols, and
that the monitor observe every action and be able to stop it.

\paragraph{The violation is the delivery of a string.}
Suppose that $\Sigma\subseteq A$ and that $A$ contains a symbol $\dlv$ that is
not in $\Sigma$.  The action $\dlv$ \emph{delivers} the string of symbols of
$\Sigma$ written since the last action that was not in $\Sigma$, or since the
start if there was none: it hands that string to a recipient who reads it, or
to a machine that runs it.  Let $B$ be a policy on delivered strings, and let
$\Bad_B$ be the set of finite traces in which a string in $B$ has been
delivered.

\begin{proposition}\label{app:full:prop:delivery}
$B\mred\Bad_B$.  Hence if $B$ is undecidable, no monitor is exact with respect
to $\Bad_B$.
\end{proposition}

\begin{proof}
For every string $w$, the trace $w\dlv$ is in $\Bad_B$ if and only if
$w\in B$.
\end{proof}

In this case the monitor must decide, before a string is delivered, whether
the string is in $B$.  Some harmful programs perform only operations that are
individually ordinary, such as writing a file, and the harm is in what is
written.  If the violation is not the delivery but a later action of whoever
receives the string, and that action is observed by a monitor, the first case
applies to that action.


\subsection{Limits of the results}\label{app:full:sec:limits}

\paragraph{Average error.}
Theorem~\ref{app:full:thm:general}(b) counts strings.  It gives no positive lower bound
on the average error of a guard under a probability distribution on outputs.
Theorem~\ref{thm:further}(b) is a different statement: it bounds the error of a randomised guard
on one particular output.  Let $P$ be a probability distribution on $\Str$ and
let
$\delta>0$.  There is a finite set $U$ of strings with $P(U)>1-\delta$.  The
guard that passes the permitted members of $U$ and blocks every other string
is sound, and the probability that it blocks a permitted output drawn from $P$
is less than $\delta$.  This holds for every policy.  If the policy is
undecidable, no single algorithm produces such a guard from every finite set
$U$, since applied to $U=\{w\}$ it would determine whether $w$ is permitted.

\paragraph{Adversaries.}
A small average error under $P$ does not bound the loss when outputs
are chosen by an adversary.  Suppose that a use of the generator is made by an
adversary with probability $\alpha$; that the adversary then obtains, with
probability $\sigma$, a forbidden output that the guard passes; that the loss
from such an output is at least $\lambda$; and that no loss is negative.  Then
the expected loss in one use is at least $\alpha\,\sigma\,\lambda$, whatever
the average error under $P$.  These quantities are assumed and are not
derived from Definition~\ref{app:full:def:guard}, and the bound holds for decidable
policies also.  If the guard is sound then $\sigma=0$; that $\sigma=0$ for a
particular adversary does not imply that the guard is sound.

\paragraph{Restricted programs.}
Suitable restrictions on the programs that a generator emits can restore
decidability.  If every program emitted halts within a number of steps
computable from its text, membership in $\Harm_H$ is decidable for those
programs by running them.  A restriction on the format of programs that does
not limit what they compute does not have this effect.

\paragraph{Summary.}
If every string is a possible output, an exact guard is a decision procedure
for the set of forbidden outputs.  There is none for a policy that forbids
every false sentence of arithmetic and no true one; for a behavioral policy
on programs that forbids some programs and permits others; or for a policy
that forbids the messages that lead an instructable recipient to a harmful
action.  The same holds for a monitor that must decide, before a string is
delivered, whether the string is forbidden by such a policy.  The results do
not exclude exact guards for outputs of bounded length or for suitably
restricted programs, or exact monitors for policies on observed actions, and
they do not bound average error under an arbitrary probability distribution on
outputs.

\end{document}